\documentclass{article} 
\let\ORIGaddcontentsline\addcontentsline
\usepackage{iclr2026_conference,times}
\let\addcontentsline\ORIGaddcontentsline

\usepackage{amsmath,amsfonts,bm}

\def\eqref#1{equation~\ref{#1}}

\def\1{\bm{1}}

\DeclareMathAlphabet{\mathsfit}{\encodingdefault}{\sfdefault}{m}{sl}
\SetMathAlphabet{\mathsfit}{bold}{\encodingdefault}{\sfdefault}{bx}{n}

\usepackage{hyperref}
\usepackage{url}
\usepackage{capt-of}
\usepackage[utf8]{inputenc} 
\usepackage[T1]{fontenc}    
\usepackage{url}            
\usepackage{booktabs}       
\usepackage{amsfonts}       
\usepackage{nicefrac}       
\usepackage{microtype}      
\usepackage{amsmath}
\usepackage{multirow}
\usepackage{subcaption}
\usepackage{pifont}
\usepackage{algorithm}
\usepackage{bbm}
\usepackage{titletoc}

\usepackage{textcomp}
\usepackage{stfloats}
\usepackage{mathabx}
\usepackage{amsthm}
\usepackage{enumitem}
\setlist{leftmargin=5mm}
\usepackage{wrapfig}
\usepackage{makecell}
\usepackage{wrapfig}
\usepackage{lipsum}
\usepackage{amsmath, amssymb, graphicx, xcolor, xspace}
\usepackage{booktabs}
\usepackage{algpseudocode}
\usepackage{bbm}
\usepackage{tcolorbox}  
\usepackage{float}
\tcbuselibrary{breakable} 
\usepackage[utf8]{inputenc}
\usepackage{tcolorbox}
\usepackage{microtype}
\newtcolorbox{mybox}[1][]{
    title=#1,
    fonttitle=\small,
    fontupper=\small,
    left=1mm,
    right=1mm,
    top=1mm,
    bottom=0mm,
}
\definecolor{darkblue}{rgb}{0, 0, 0.5}
\definecolor{darkred}{rgb}{0.72, 0.22, 0.27}
\definecolor{lightblue}{RGB}{129, 209, 241}
\definecolor{forestgreen}{RGB}{34, 139, 34}

\usepackage{inconsolata}

\usepackage{graphicx}
\usepackage[utf8]{inputenc} 
\usepackage[T1]{fontenc}    
\usepackage{url}            
\usepackage{booktabs}       
\usepackage{amsfonts}       
\usepackage{nicefrac}       
\usepackage{microtype}      
\usepackage{xcolor}         
\usepackage{amssymb}
\usepackage{amsmath}
\usepackage{multirow}
\usepackage{subcaption}
\usepackage{amssymb}
\usepackage{algorithm}
\usepackage{bbm}
\usepackage{stfloats}
\usepackage{mathabx}
\usepackage{amsthm}
\usepackage[table]{xcolor}
\usepackage{enumitem}
\setlist{leftmargin=5mm}
\usepackage{graphicx}
\usepackage{wrapfig}
\usepackage{fontawesome5}
\usepackage{makecell}
\usepackage[dvipsnames]{xcolor}
\usepackage{wrapfig}
\usepackage{lipsum}
\usepackage{amsmath, amssymb, graphicx, xcolor, xspace}
\usepackage{booktabs}
\usepackage{algpseudocode}
\usepackage{bbm}
\usepackage{listings}
\definecolor{codekw}{HTML}{9B2393}   
\definecolor{codecls}{HTML}{0B7285}  
\definecolor{codestr}{HTML}{C0392B}  
\definecolor{codecmt}{HTML}{6B7280}  
\definecolor{codebg}{HTML}{F5F6F7}   
\lstdefinestyle{llmrouter}{
  language=Python,
  basicstyle=\ttfamily\small,
  keywordstyle=\color{codekw}\bfseries,
  stringstyle=\color{codestr},
  commentstyle=\color{codecmt}\itshape,
  emph={MetaRouter,BaseTrainer,MyRouter,MyRouterTrainer,nn},
  emphstyle=\color{codecls},
  morekeywords={self},
  showstringspaces=false,
  breaklines=true,
  columns=fullflexible,
  keepspaces=true,
  xleftmargin=6pt,xrightmargin=6pt,aboveskip=4pt,belowskip=2pt,
  backgroundcolor=\color{codebg},
  frame=none,
}

\definecolor{gMMAU}{HTML}{C0D9ED}
\definecolor{gMMAR}{HTML}{F9E8C5}
\definecolor{gMMSU}{HTML}{CFBAD9}
\definecolor{gOVR}{HTML}{CFBAD9}
\usepackage{tikz}
\makeatletter
\newcommand{\DrawLine}{%
  \begin{tikzpicture}
  \path[use as bounding box] (0,0) -- (\linewidth,0);
  \draw[color=black,dashed,dash phase=2pt]
        (0-\kvtcb@leftlower-\kvtcb@boxsep,0)--
        (\linewidth+\kvtcb@rightlower+\kvtcb@boxsep,0);
  \end{tikzpicture}%
  }

\definecolor{violet}{RGB}{138, 43, 226}

\makeatother

\definecolor{citepcol}{HTML}{2DDC0E}
\definecolor{tableofcontent}{HTML}{E63E15}
\definecolor{urlcol}{HTML}{2470D8}
\definecolor{myorange}{RGB}{2, 142, 2}
\hypersetup{
    colorlinks=true,
    citecolor=darkblue,
    linkcolor=red,
    filecolor=magenta,      
    urlcolor=cyan,
}
\def\project{\raisebox{-0.5pt}{\textcolor{black}{\faGlobe}}}

\definecolor{barGray}{HTML}{D9D9D9} 
\NewDocumentCommand{\shibo}{ mO{} }
{\textcolor{pink}{\textsuperscript{\textit{Shibo}}\textsf{\textbf{\small[#1]}}}}

\usepackage{xcolor}

\definecolor{RouteCyan}{HTML}{0EA5E9}
\definecolor{RouteViolet}{HTML}{7C3AED}

\definecolor{barGray}{HTML}{D9D9D9} 

\ExplSyntaxOn
\NewDocumentCommand{\gradienttext}{mmm}
 {
  \llmr_gradient:nnn { #1 } { #2 } { #3 }
 }
\cs_new_protected:Nn \llmr_gradient:nnn
 {
  \tl_set:Nn \l_tmpa_tl { #1 }
  \int_zero:N \l_tmpa_int
  \int_set:Nn \l_tmpb_int { \tl_count:N \l_tmpa_tl }
  \tl_map_inline:Nn \l_tmpa_tl
   {
    \int_incr:N \l_tmpa_int
    \fp_set:Nn \l_tmpa_fp { 100 * (\l_tmpa_int - 1) / (\l_tmpb_int - 1) }
    \textcolor{ #3 ! \fp_to_decimal:N \l_tmpa_fp ! #2 }{ ##1 }
   }
 }
\ExplSyntaxOff

\title{\raisebox{-0.10\height}{\includegraphics[height=1.1em]{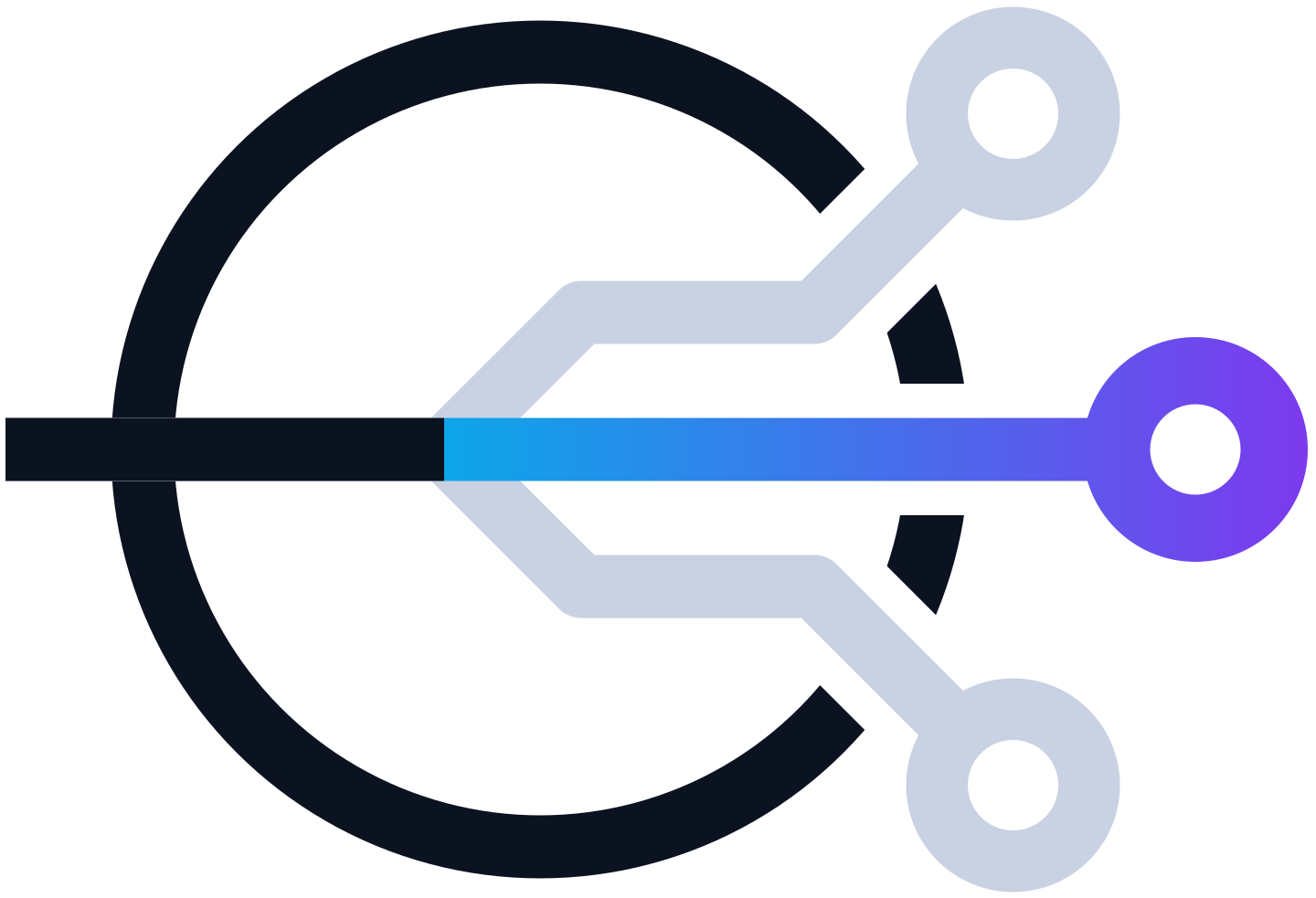}} \texorpdfstring{\scalebox{1.25}{\gradienttext{LLMRouter}{RouteCyan}{RouteViolet}}}{LLMRouter}: Unified Infrastructure for Developing, Evaluating, and Deploying LLM Routers}

\author{%
Tao Feng\textsuperscript{1*},\;\;
Fangxu Yu\textsuperscript{2*},\;\;
Haozhen Zhang\textsuperscript{3*},\;\;
Zhongjie Dai\textsuperscript{1},\;\;
Liangqi Yuan\textsuperscript{4},\;\;
Zijie Lei\textsuperscript{1},\;\;\\
\textbf{Weizhi Zhang}\textsuperscript{\textbf{5}},\;\;
\textbf{Kunlun Zhu}\textsuperscript{\textbf{1}},\;\;
\textbf{Haodong Yue}\textsuperscript{\textbf{1}},\;\;
\textbf{Keyang Xuan}\textsuperscript{\textbf{1}},\;\;
\textbf{Ge Liu}\textsuperscript{\textbf{1}},\;\;
\textbf{Jiaxuan You}\textsuperscript{\textbf{1}}
\\
\textsuperscript{1}University of Illinois Urbana-Champaign,\;\;
\textsuperscript{2}University of Maryland, College Park,\\
\textsuperscript{3}Nanyang Technological University,\;\;
\textsuperscript{4}Purdue University,\;\;
\textsuperscript{5}University of Illinois Chicago\\
\\
\makebox[\linewidth][c]{%
\project\;\href{https://ulab-uiuc.github.io/LLMRouter/}{Project}\hspace{2em}
\huggingface\;\href{https://huggingface.co/datasets/ulab-ai/xRouteBench}{xRouteBench}\hspace{2em}
\github\;\href{https://github.com/ulab-uiuc/LLMRouter}{Code}\hspace{3em}}\\
}

\newcommand{\method}{\texttt{LLMRouter}\xspace}
\def\huggingface{\raisebox{-1.5pt}{\includegraphics[height=1.05em]{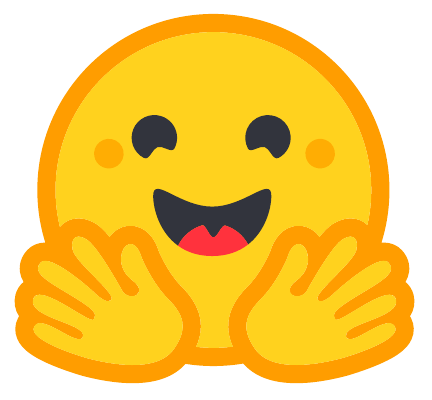}}}
\def\github{\raisebox{-1.5pt}{\includegraphics[height=1.0em]{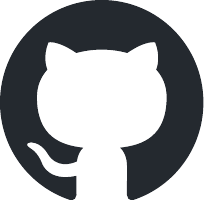}}}

\newcommand{\msheader}{\raisebox{-3pt}{\includegraphics[height=14pt]{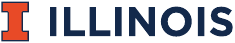}}}
\fancypagestyle{msfirst}{%
  \fancyhf{}%
  \fancyhead[L]{\msheader}%
  \fancyhead[R]{}%
  \renewcommand{\headrulewidth}{0.8pt}%
}

\iclrfinalcopy 
\begin{document}

\maketitle
\thispagestyle{msfirst}
{\let\thefootnote\relax\footnotetext{\textsuperscript{*}Equal contribution.}}

\begin{abstract}
No single large language model (LLM) is optimal across all queries and budget constraints, making model routing essential for cost-effective LLM deployment.
Existing routers span binary quality predictors, cost-aware cascades, graph-based routers, and agentic routers, yet their diverse formalisms and incompatible implementations, coupled with the absence of a standardized evaluation pipeline, hinder fair comparison and further extension.
In this paper, we present a unified formulation of LLM routing as a sequential decision process. Under this formulation, a router can be characterized in terms of five types of components: context encoders, model encoders, scoring functions, decision rules, and learning signals. Existing methods can then be organized into three families of single-turn, multi-turn, and personalized routing.
Building on this formulation, we develop an automated pipeline that constructs routing supervision by systematically running a pool of candidate models across benchmarks and evaluates routers in terms of both response quality and inference cost under a unified protocol. The resulting benchmark, \textbf{xRouteBench}, spans generic LLM tasks, memory-augmented, vision (image and video), time-series, and personalized routing scenarios.
Grounded in the formulation and pipeline, we present \textbf{\method}, an open-source infrastructure for standardized and modular implementation of LLM routers, where users can add a new router by implementing only a routing method and a loss function and access built-in implementations of more than 16 representative routers spanning all three families.
Using the library and benchmark, we conduct a systematic empirical study of LLM routing and find that learned routers achieve a 14.6\% relative improvement over the strongest fixed-model baseline, router rankings reverse in favor of lightweight designs under tighter cost constraints, and user-conditioned routing delivers consistent personalization gains.
\end{abstract}

\section{Introduction}
\label{sec:intro}

The rapid proliferation of large language models (LLMs) has created a heterogeneous ecosystem of models with widely varying costs and task-specific capabilities, ranging from frontier systems to substantially cheaper open-weight alternatives. Since no single model is optimal across all queries and budget constraints, model routing, which determines which model should handle each query, has become essential for cost-effective LLM deployment. Beyond cost efficiency, routing also matches each query to the candidate model best suited to it and adapts model choice to user-specific preferences (Figure~\ref{fig:main_arch}). 
Rich research has been devoted to this problem, from binary routers that arbitrate between a weak and a strong model \citep{ding2024hybrid, ong2024routellm} and cost-aware cascades \citep{chen2023frugalgpt, aggarwal2024automix}, to reward-guided ensembles, contrastive and graph-based routers \citep{chen2024routerdc, feng2024graphrouter}, personalized routers that adapt to individual users \citep{xie2025gmtrouter, dai2025personalizedrouter}, and agentic routers trained with reinforcement learning \citep{zhang2025router, feng2026graphplanner}.

Despite this rapid progress, the field still lacks a unified foundation on which these diverse approaches can be developed and compared, mainly due to two obstacles.
First, existing routers are developed under distinct formalisms, released as separate codebases with incompatible interfaces, trained with different supervision, and tuned for different candidate pools, making it difficult to isolate the design elements that actually drive performance or determine whether observed differences stem from the routers themselves or from their broader experimental stacks.
Second, evaluating a router is fundamentally more demanding than evaluating a single model, as constructing routing supervision and enabling standardized evaluation require running every candidate model on every benchmark query and scoring each response with task-specific metrics. Existing benchmarks precompute candidate responses for fixed model pools \citep{hu2024routerbench, huang2025routereval}, but are limited to single-turn routing and provide no pipeline for generating supervision for new benchmarks or candidate pools. Consequently, multi-turn and personalized routing still lack a standardized, cost-aware evaluation framework, making routing methods difficult to compare, reuse, and transfer from offline studies to real applications.

In this paper, we introduce \textbf{\method}, a unified infrastructure for developing, evaluating, and deploying routing policies over heterogeneous LLM backends. Within this infrastructure, a router is characterized by five types of components: context encoders, model encoders, scoring functions, decision rules, and learning signals. This abstraction accommodates existing routers, which we group into three families of single-turn, multi-turn (including agentic), and personalized routing. It also reduces the effort to add a new router to implementing a routing method and a loss function, while data construction, training, inference, and evaluation apply unchanged, so switching the router, candidate pool, or training objective requires only a configuration change rather than reimplementation. \method\ includes more than 16 representative routers spanning all three families, and can expose any of them as an OpenAI-compatible server for deployment on messaging platforms via OpenClaw~\citep{openclaw2026} or through a ComfyUI-based visual interface for code-free prototyping.

\method\ further automates the construction of routing supervision and evaluation, the main obstacle to comparing routers on equal footing. Its pipeline assembles queries from established benchmarks, dispatches each query to a pool of 18 candidate models spanning a broad price range, and scores every response with task-specific metrics while recording token-level cost. Every router is then evaluated on the same queries, candidate pool, and metrics, enabling direct comparison of their quality-cost trade-offs. With this pipeline, we construct \textbf{xRouteBench}, a benchmark that spans generic LLM tasks, memory-augmented, vision (image and video), time-series, and personalized scenarios under one protocol.



Leveraging \method\ and xRouteBench, we conduct a systematic empirical study of LLM routing across the three families under one protocol. Our study surfaces four findings:
(i) \textbf{no single router dominates}, as the best router varies across tasks and cost budgets. Strong average performance therefore reflects consistency across scenarios rather than dominance in any single setting. (ii) \textbf{learned routing still outperforms the strongest fixed-model baseline}, because always selecting the largest model incurs the highest cost yet delivers only mediocre performance, whereas learned routers select smaller, cheaper models for many queries that the largest model answers incorrectly. (iii) \textbf{multi-turn routing does not consistently outperform single-turn routing}, as additional rounds of decomposition and aggregation often add cost and redundant information, and their benefit hinges on the capability of the base model that performs them. (iv) \textbf{personalization pays off, but only when user context is modeled well}, as a user-conditioned router ranks first under both the persona judge and real human preferences, yet the two settings favor different personalized designs.
We also release the library and benchmark in the hope of fostering more systematic progress in LLM routing.

\section{A Unified Formulation of LLM Routing}
\label{sec:formulation}

\subsection{Routing as a Sequential Decision Process}
\label{sec:general-router}
\begin{figure*}[t]
\centering
\includegraphics[width=1.0\textwidth]{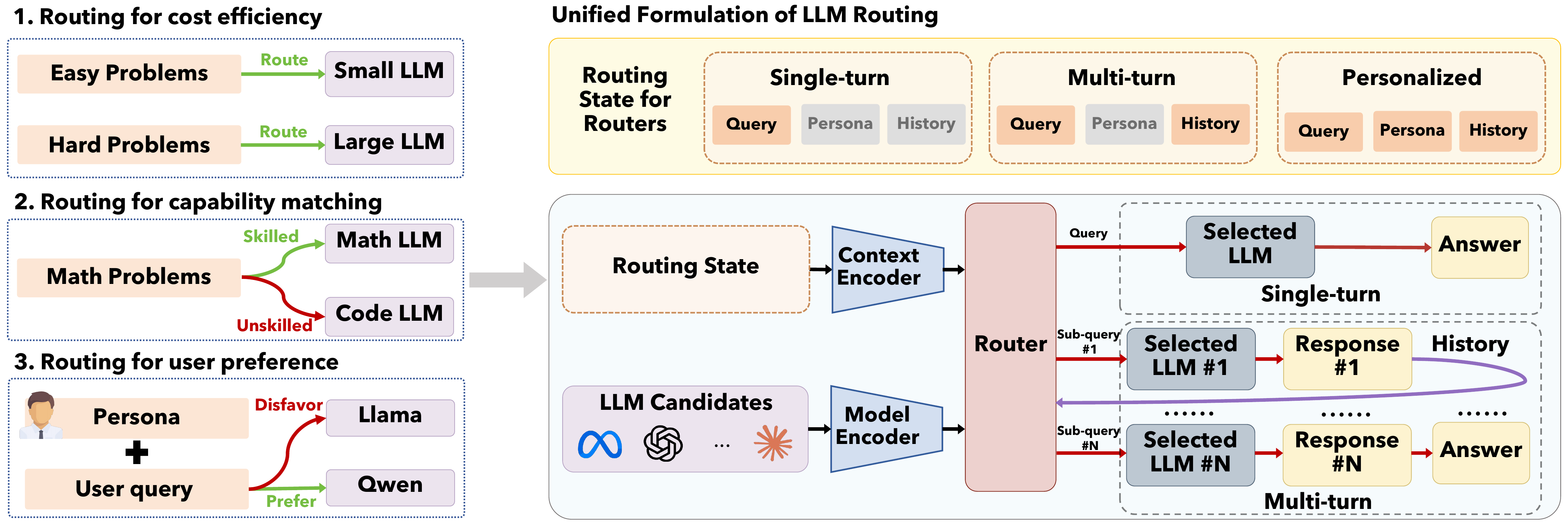}
\caption{\textbf{Overview of LLM routing.} Routing is driven by three needs (left), namely cost efficiency, capability matching, and user preference. Our unified formulation (right) casts all of them as one decision process: a context encoder $E_q$ represents the routing state of query, persona, and interaction history, a model encoder $E_m$ represents each candidate, and the router dispatches the query or its sub-queries to selected models and aggregates their responses into the answer. The single-turn, multi-turn, and personalized families differ only in which part of the state they observe.}
\label{fig:main_arch}
\vspace{-10pt}
\end{figure*}

Existing routers take seemingly incompatible forms, from binary routers that arbitrate between a weak and a strong model \citep{ding2024hybrid, ong2024routellm} and cost-aware cascades \citep{chen2023frugalgpt, aggarwal2024automix} to graph-based routers \citep{feng2024graphrouter, yu2026tsrouter} and agentic routers trained with reinforcement learning \citep{zhang2025router}, yet they can all be formulated as a sequential decision process. At step $t$, the router observes a state $s_t = (q, u, h_t)$, consisting of the input query $q$, an optional user context $u$ (e.g., a user identifier with past interactions and feedback), and the interaction history $h_t$ accumulated so far, and takes an action $a_t \in \mathcal{M} \cup \{\bot\}$. The dispatch action $a_t = m$ sends the state to candidate $m$ from the pool $\mathcal{M} = \{m_1, \dots, m_K\}$ and appends its response $y_t$ to the history, $h_{t+1} = h_t \oplus y_t$, while the terminating action $a_t = \bot$ ends the episode and aggregates the collected responses into the final answer $y$; single-turn routing is the special case that terminates after one dispatch. The goal of routing is a policy $\pi$ whose trajectory $\tau = (a_1, \dots, a_T)$ produces high-quality answers at low inference cost:
\begin{equation}
    \pi^\star \;=\; \arg\max_{\pi}\;
    \mathbb{E}_{q,\; \tau \sim \pi}
    \big[\, \mathrm{perf}(y \mid q) \;-\; \lambda \cdot c(\tau) \,\big],
    \label{eq:objective}
\end{equation}
where $\mathrm{perf}(y \mid q)$ aggregates task-specific quality metrics (e.g., accuracy, F1, or an LLM-judged score), $c(\tau)$ sums the monetary or token cost of every call in the routing trajectory $\tau$, and $\lambda \geq 0$ controls the performance--cost trade-off. Under this formulation, a router is characterized by the choice of a \textbf{context encoder} $E_q$ that encodes the routing state, a \textbf{model encoder} $E_m$ that encodes each LLM candidate, a \textbf{scoring function} $g$ and a \textbf{decision rule} $d$ that turn the context and model representations into a routing action, and a \textbf{learning signal} $\mathcal{L}$ that fits these components toward the optimal policy. This section elaborates on each component and demonstrates how existing routers fall into three families, namely single-turn, multi-turn (including agentic), and personalized, with specific designs of these five components (Table~\ref{tab:formulation}). We provide an overview in Figure~\ref{fig:main_arch}.

\noindent\textbf{Context encoder.}
The context encoder $E_q$ maps the routing state $s_t$ to the representation on which the routing decision is based, and its output takes one of two forms. i) \textbf{Embedding-based}: the state is represented as a vector. Rating-based routers degenerate to a constant that ignores the query and routes by global model quality, $k$NN-style routers use an off-the-shelf sentence embedding \citep{hu2024routerbench, shnitzer2023large}, discriminative routers train a lightweight encoder over frozen embeddings \citep{ding2024hybrid, stripelis2024tensoropera}, and personalized routers condition the representation on user and session nodes of a heterogeneous interaction graph \citep{xie2025gmtrouter, dai2025personalizedrouter}. ii) \textbf{Text-based}: the state is kept in natural language. Cascades append the draft response and a verification confidence to the query \citep{chen2023frugalgpt, aggarwal2024automix}, and fine-tuned LM routers, exemplified by Router-R1, verbalize the whole state directly in the prompt, leaving its representation to the model's forward pass \citep{ong2024routellm, zhang2025router}. Which portion of the state $E_q$ reads is precisely what separates the three router families, and any router is personalized by swapping in a user-conditioned $E_q$ while inheriting the remaining components.

\noindent\textbf{Model encoder.}
The model encoder $E_m$ encodes each candidate in the pool. i) \textbf{Static metadata}: the simplest choice describes a candidate by its model size, capability description, and pricing. ii) \textbf{Historical profiles}: most routers instead profile candidates by their past behavior, representing a model by the set of embedded queries it has previously solved ($k$NN), a scalar rating (Elo), or a latent factor fit by matrix factorization. iii) \textbf{Learned embeddings}: stronger routers learn model embeddings jointly with the context encoder \citep{chen2024routerdc, feng2024graphrouter, zhuang2025embedllm}. iv) \textbf{Verbalized description}: fine-tuned LM routers instead name the candidates directly in the prompt \citep{ong2024routellm, zhang2025router}.

\noindent\textbf{Scoring function and decision rule.}
The scoring function $g$ measures the compatibility between the encoded state and each candidate, and the decision rule $d$ converts the resulting scores into a routing action. Instantiations of $g$ track the encoders, from embedding similarity in $k$NN-style routers and a bilinear product in factorization-based ones, to a classification head \citep{ding2024hybrid, stripelis2024tensoropera}, message passing over a query--model graph \citep{feng2024graphrouter}, and next-token logits in fine-tuned LM routers that fold $E_q$, $E_m$, and $g$ into one forward pass \citep{ong2024routellm}. For $d$, greedy $\arg\max$ is the default choice, yet it is optimal for Eq.~\ref{eq:objective} only when $\lambda = 0$. Cost-aware rules instead threshold the predicted quality gap between a cheap and an expensive model \citep{ding2024hybrid}, accept or escalate in cascades \citep{chen2023frugalgpt, aggarwal2024automix}, or sample for exploration in online settings \citep{dai2024cost}, and multi-turn routers further equip $d$ with the terminating action $\bot$ \citep{zhang2025router}.

\noindent\textbf{Learning signal.}
The learning signal $\mathcal{L}$ specifies how the components above are fit toward Eq.~\ref{eq:objective}. Non-parametric routers require no training and rely purely on stored interactions. Supervised routers fit pointwise correctness labels harvested by running the candidate pool over benchmark queries, preference-based routers learn from pairwise comparisons such as human votes from Chatbot Arena \citep{ong2024routellm} or contrastive objectives that pull queries toward the models that solve them \citep{chen2024routerdc}, and agentic routers directly optimize trajectory-level rewards with reinforcement learning \citep{zhang2025router}. In every case, $\mathcal{L}$ is a surrogate for the same objective. What differs is not the goal but the form in which $\mathrm{perf}$ is observable, measured for every candidate by supervised routers, returned only at the end of a trajectory for agentic ones, and revealed only through comparisons when quality is user-specific.

\begin{table}[t]
\centering
\caption{\textbf{Instantiation of the unified routing formulation for the three router families.} For each family, the table specifies the routing state $s$, the context and model encoders $E_q$ and $E_m$, the routing action defined by the scoring function $g$ and decision rule $d$, and the learning signal $\mathcal{L}$ used to optimize response quality and inference cost.}
\label{tab:formulation}
\small
\setlength{\tabcolsep}{4pt}
\resizebox{\linewidth}{!}{%
\begin{tabular}{lllll}
\toprule
\textbf{Family} & \textbf{State} $s$ & \textbf{Encoders} $E_q, E_m$ & \textbf{Routing action} (scoring $g$, decision $d$) & \textbf{Learning signal} $\mathcal{L}$ (surrogate of Eq.~\ref{eq:objective}) \\
\midrule
Single-turn & $(q)$ & $E_q(q),\; E_m(m)$ & $a = \arg\max_{m \in \mathcal{M}}\, g\big(E_q(q),\, E_m(m)\big)$ & fit $g$ to per-candidate reward $\mathrm{perf}(y_m \mid q) - \lambda\, c_m$ \\
\midrule
Multi-turn & $(q, h_t)$ & $E_q(q, h_t),\; E_m(m)$ & $a_t \sim d\big(\{ g(E_q(q, h_t),\, E_m(m)) \}_{m}\big)$ & maximize episode return $\mathbb{E}_{\tau}\big[\mathrm{perf}(y \mid q) - \lambda\, c(\tau)\big]$ \\
\midrule
Personalized & $(q, u, h_t)$ & $E_q(q, u, h_t),\; E_m(m)$ & $a = \arg\max_{m \in \mathcal{M}}\, g\big(E_q(q, u, h_t),\, E_m(m)\big)$ & fit $g$ to comparisons $m^+ \!\succ_u\! m^-$ observing $\mathrm{perf}_u$ \\
\bottomrule
\end{tabular}%
}
\vspace{-10pt}
\end{table}

\subsection{Automatic Evaluation of LLM Routing}
\label{sec:protocol}
Evaluating a router is substantially more demanding than evaluating a single model. Under Eq.~\ref{eq:objective}, a router must be judged on both the quality of its answers and the cost spent to obtain them, and constructing its supervision requires knowing how every candidate performs on every query under task-specific metrics. In current practice, these elements are assembled manually for a single benchmark and fixed candidate pool \citep{hu2024routerbench, huang2025routereval}, requiring fresh engineering for every new task or candidate pool.

\method\ automates this process end-to-end with a three-stage pipeline: i) \textbf{Query Curation}: queries are sampled from source benchmarks, normalized into a unified schema, and split into training and test sets; ii) \textbf{Response Collection}: each query is dispatched to every candidate in the pool, which is declared in a single configuration file, and responses are collected together with their token counts; iii) \textbf{Metric Scoring and Pricing}: every response is scored with its task metric and priced from its token counts. The product is a dense query--model matrix of performance and cost that serves at once as routing supervision and as the test bed, so a new task or candidate pool enters through a configuration change rather than a re-engineered stack.

Evaluation reuses this path, except that a test query goes only to the candidate the router selects rather than to the whole pool. Every router therefore faces the same queries, candidate pool, and metrics, so measured differences reflect the routing policy rather than the surrounding stack. For the multi-turn and agentic families, every decomposition and aggregation call is priced into the trajectory cost.

\noindent\textbf{Metrics.} Task quality is assessed using built-in metrics aligned with standard benchmark conventions, including exact and close matching, multiple-choice accuracy, token-level F1, mathematical answer verification, and execution-based code evaluation. \method\ also supports optional LLM-based judging and exposes a weighted objective that balances performance and cost, allowing routers and trainers to target performance-first, cost-sensitive, or hybrid operating points.

\section{xRouteBench: A Multi-Scenario Benchmark for LLM Routing}
\label{sec:xroutebench}
Existing routing benchmarks cover only a subset of the settings captured by the formulation in \S\ref{sec:general-router}. RouterBench \citep{hu2024routerbench} precomputes candidate responses for single-turn text queries over a fixed candidate pool, but does not cover settings in which input-token costs dominate or only a subset of candidate models can process the input. RouterEval \citep{huang2025routereval} aggregates large-scale performance records but likewise focuses on single-turn text tasks and evaluates response quality independently of inference cost. Recent vision--language routing benchmarks \citep{huang2025vl, ma2026mmr} extend routing evaluation to image inputs, but remain limited to image question answering and do not cover video, long-context, or modality-selection settings. Preference data from Chatbot Arena \citep{zheng2023judging} provides population-level signals but lacks the persistent user context needed to supervise user-conditioned routing. To close these gaps, we construct xRouteBench to evaluate, under a unified cost-aware protocol, regimes in which routing decisions fundamentally differ: long-context inputs for which input-token costs dominate, image and video inputs that only a subset of candidate models can process, time-series inputs with multiple modality encodings, and tasks with user-specific quality preferences.

\begin{wrapfigure}{r}{0.5\textwidth}
\centering
\vspace{-15pt}
\includegraphics[width=0.5\textwidth]{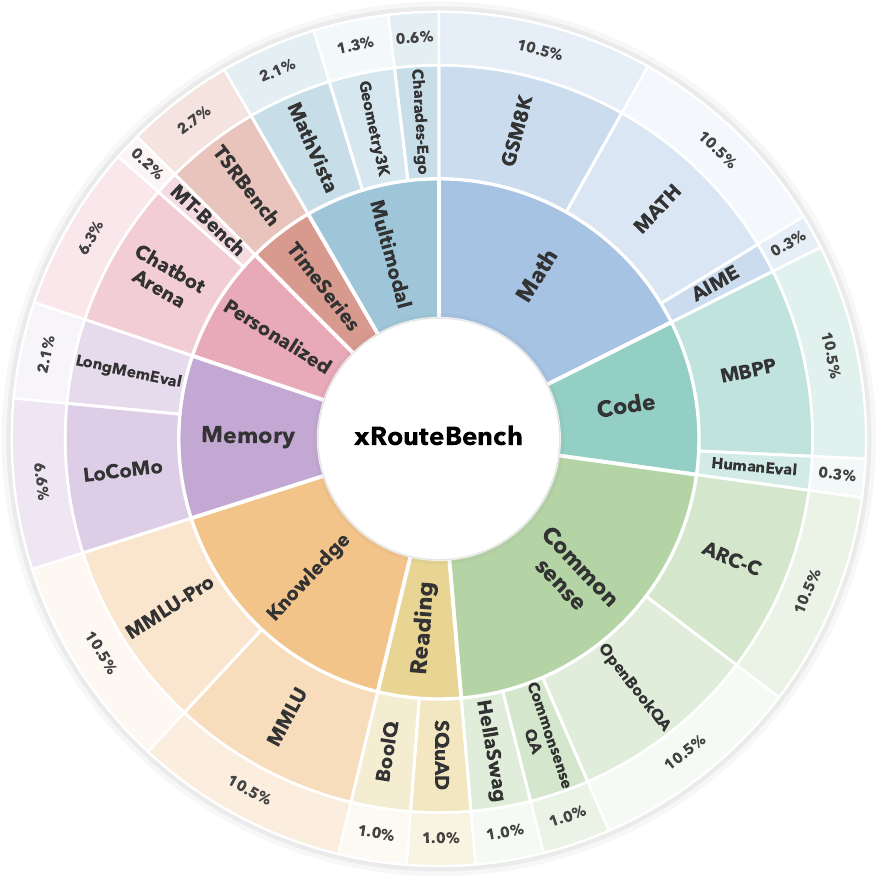}
\caption{\textbf{Task composition of xRouteBench.} The benchmark covers generic LLM tasks, memory, vision, time-series, and personalized routing, with percentages indicating the proportion of test queries contributed by each dataset.}
\label{fig: xroutebench}
\vspace{-15pt}
\end{wrapfigure}

\noindent\textbf{Design principle.} All tasks are constructed by the \method\ data engine and share a common query schema, supervision format, and evaluation protocol. Each non-text asset is converted by a transformation script into a self-contained textual query, with an optional pointer to the source image, video, or time series. This separates routing from perception by ensuring that text-only and multimodal candidates receive the same textual input. For every query, the protocol jointly evaluates response quality and inference cost, exposing regimes in which input-token costs dominate answer-generation costs. Adding a new application requires only a transformation script and a registered metric.

\noindent\textbf{Tracks.} xRouteBench spans five tracks comprising 4{,}767 instances. Figure~\ref{fig: xroutebench} provides an overview of task distribution. Specifically, we include: 
(i) \textit{Generic LLM Tasks} mix established knowledge and commonsense QA (MMLU \citep{hendrycks2020measuring}, MMLU-Pro \citep{wang2024mmlupro}, ARC-Challenge \citep{allenai:arc}, OpenBookQA \citep{mihaylov2018can}, CommonsenseQA \citep{talmor2019commonsenseqa}, BoolQ \citep{clark2019boolq}, HellaSwag \citep{zellers2019hellaswag}, SQuAD \citep{rajpurkar2016squad}), mathematical reasoning (GSM8K \citep{cobbe2021training}, MATH \citep{hendrycks2021measuring}, AIME), and code generation (MBPP \citep{austin2021program}, HumanEval \citep{chen2021evaluating}), the conventional single-turn text setting.
(ii) \textit{Memory} routes long-horizon conversational QA over hundreds of accumulated turns (LoCoMo \citep{maharana2024evaluating}, LongMemEval \citep{wu2024longmemeval}), where token cost is governed by the history rather than the answer.
(iii) \textit{Vision} covers image-grounded mathematical reasoning (Geometry3K \citep{lu2021inter}, MathVista \citep{lu2024mathvista}) and egocentric video understanding (Charades-Ego \citep{sigurdsson2018charades}).
(iv) \textit{TimeSeries} covers time-series reasoning (TSRBench \citep{yu2026tsrbench}), with each series rendered as both text and image so the router also selects a modality encoding.
(v) \textit{Personalized} draws open-ended prompts from Chatbot Arena and MT-Bench~\citep{zheng2023judging}, each tied to a user persona and scored by a persona-conditioned LLM judge, so its supervision is preference feedback rather than pointwise correctness.


\section{The \method\ Library}
\label{sec:library-design}
\begin{figure*}[t]
\centering
\includegraphics[width=1.0\textwidth]{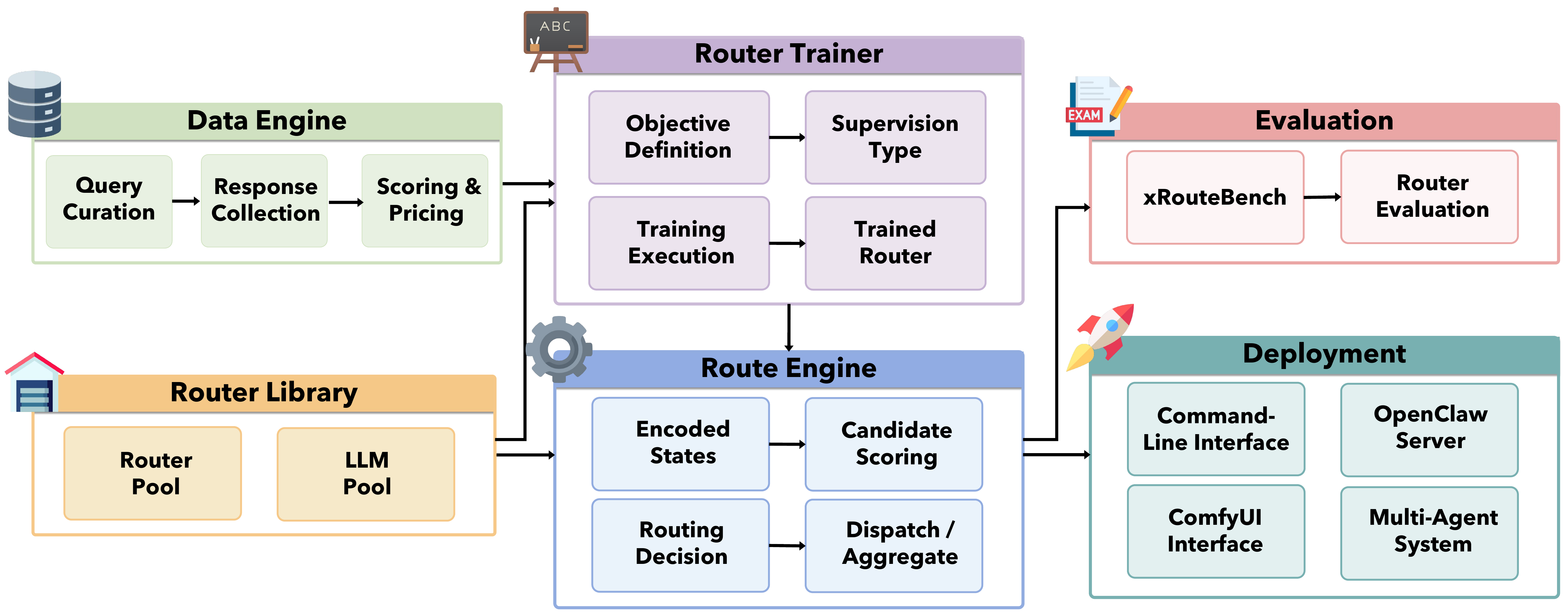}
\caption{\textbf{Architecture of \method.} The system consists of six modules that support routing data construction, router implementation and training, inference, evaluation, and deployment.  
}
\label{fig:library}
\vspace{-15pt}
\end{figure*}
\method\ ties the formulation of \S\ref{sec:general-router}, the evaluation protocol of \S\ref{sec:protocol}, and xRouteBench into one executable system organized as six modules around a single query--model matrix (Figure~\ref{fig:library}). Its organizing principle is that the five components of a router are the only thing a user writes, while data construction, training, inference, evaluation, and deployment are shared infrastructure that operates on any router unchanged. Swapping a router, a candidate pool, or a training objective is therefore a configuration change rather than a reimplementation.

\noindent\textbf{Data Engine.} The data engine implements the three-stage construction pipeline of \S\ref{sec:protocol}, turning a declared task list and candidate pool into the query--model matrix that supervises and tests every router. Adding a task requires only a prompt template and a registered metric, and adding a candidate requires only its endpoint and per-token price.

\noindent\textbf{Router Library.} The library implements more than 16 routers spanning all three families under a unified \texttt{MetaRouter} interface, which hides mechanisms as different as nearest-neighbor retrieval in a $k$NN router and autoregressive decoding in a fine-tuned LM router behind one call. Adding a new router requires subclassing \texttt{MetaRouter} and implementing either \texttt{route\_single} or its batched counterpart, \texttt{route\_batch}. Within this method, the context encoder $E_q$, model encoder $E_m$, scoring function $g$, and decision rule $d$ map a routing state to a routing action. A short YAML file specifies the router's candidate pool and objective weights, after which the router can be invoked by name using the same commands as any built-in method. Personalization and component ablations can then be performed with a one-line change, without forking the codebase. Figure~\ref{fig:code} shows the complete code needed to design a new router.



\begin{figure}[t]
\centering
\begin{lstlisting}[style=llmrouter]
from llmrouter.models import MetaRouter, BaseTrainer

class MyRouter(MetaRouter):                    # (E_q, E_m, g, d): state -> action
    def route_single(self, query):
        s = self.encode_state(query)           # context encoder  E_q
        scores = self.score(s, self.models)    # model encoder E_m + scoring g
        query["model_name"] = self.decide(scores)   # decision rule d
        return query

class MyRouterTrainer(BaseTrainer):            # learning signal  L
    def loss_func(self, outputs, batch):
        return my_objective(outputs, batch)    # pointwise / pairwise / RL reward

# Train and run through the same interface as every built-in router.
router  = MyRouter(yaml_path="my_router.yaml")
trainer = MyRouterTrainer(router)
trainer.train()
answer  = router.route_single({"query": "..."})
\end{lstlisting}
\caption{\textbf{The five components of the routing formulation map onto two classes in \method}. A router subclasses \texttt{MetaRouter} and implements \texttt{route\_single} (or \texttt{route\_batch}), where the context encoder $E_q$, model encoder $E_m$, scoring function $g$, and decision rule $d$ turn a state into a selected model; the learning signal $\mathcal{L}$ lives in a \texttt{BaseTrainer} subclass. }
\label{fig:code}
\vspace{-10pt}
\end{figure}

\noindent\textbf{Trainer.} Training is decoupled from routing through a \texttt{BaseTrainer}. Its \texttt{loss\_func} defines the learning signal $\mathcal{L}$ as a pointwise loss, pairwise loss, or trajectory-level reward, while its \texttt{train} loop uses this signal to optimize the router for the weighted objective in Eq.~\ref{eq:objective}. A router and its trainer are paired but can be swapped independently, allowing the same scorer to be trained with different forms of supervision without modifying its routing code. Non-parametric routers bypass this module.

\noindent\textbf{Route Engine.} At inference time, the route engine drives any router through the same call, dispatching the query to the selected candidate. For multi-turn policies, it repeats the decision step until a termination action is produced and aggregates the collected responses into the final answer.

\noindent\textbf{Evaluation.} The evaluation module implements the evaluation protocol of \S\ref{sec:protocol}, scoring each router on the same test queries, candidate pool, and metrics, and sweeping the trade-off weight $\lambda$ to trace its performance--cost frontier.

\noindent\textbf{Deployment.} In addition to training and inference commands, \method\ can expose any router as an OpenAI-compatible server that integrates with OpenClaw~\citep{openclaw2026} for deployment on messaging platforms such as Slack and Discord. A routing memory persists the interaction history $h$ across turns, while a ComfyUI-based canvas supports code-free prototyping. The same router evaluated offline can therefore serve live single-agent and multi-agent traffic without modification.

\section{Experiments} \label{sec:exp}
\subsection{Experimental Setups}
\noindent\textbf{Benchmarks.}
We evaluate routers across the five xRouteBench tracks defined in \S\ref{sec:xroutebench}: Generic LLM Tasks, memory, vision, time-series, and personalized. Together, they comprise eight test sets, with full statistics reported in Table~\ref{tab:app-benchmarks}. For each query in the memory track, we retrieve up to five memory items as context and score responses using token-level F1.

\noindent\textbf{LLM Candidates.}
The candidate pool contains 18 open-weight models served through two providers (i.e., Together API\footnote{\url{https://www.together.ai/}} and NVIDIA NIM API\footnote{\url{https://build.nvidia.com/}}), spanning 7B to 671B parameters. It covers Gemma-2-9B \citep{team2024gemma}; Mistral-7B, Mistral-Small-24B, Mixtral-8x7B, and Mixtral-8x22B \citep{jiang2023mistral, jiang2024mixtral}; Qwen2.5-7B, Qwen3-Next-80B, and Qwen3-Coder \citep{yang2024qwen2, yang2025qwen3}; Llama-3-8B, Llama-3-70B, Llama-3.3-70B, and Llama-4-Maverick \citep{grattafiori2024llama, adcock2026llama}; GPT-OSS-20B and GPT-OSS-120B \citep{agarwal2025gpt}; RNJ-1-15B \citep{rnj1_5_instruct}; and the two 671B models DeepSeek-V3.1 \citep{liu2024deepseek} and Cogito-v2 \citep{deepcogito2025cogitov2_671b}. Per-token pricing is given in Table~\ref{tab:app-pricing} in the appendix.

\noindent\textbf{Implemented Routers.}
\method\ implements more than 16 routers covering three families. (i) \textbf{Single-turn routers} include $k$NNRouter~\citep{li2025rethinking}, SVMRouter, MLPRouter, EloRouter, and MFRouter~\citep{ong2024routellm, shnitzer2023large}, as well as RouterDC \citep{chen2024routerdc}, Hybrid LLM \citep{ding2024hybrid}, AutoMix \citep{aggarwal2024automix}, GraphRouter \citep{feng2024graphrouter}, CausalLM~\citep{ong2024routellm}, and two rule-based baselines that always select the smallest or largest model; (ii) \textbf{Multi-turn routers} include Router-R1 \citep{zhang2025router} together with $k$NN-based and LLM-based multi-round routers; and (iii) \textbf{Personalized routers} include GMTRouter~\citep{xie2025gmtrouter} and PersonalizedRouter~\citep{dai2025personalizedrouter}.

\noindent\textbf{Evaluation Protocol.}
We score each router by a weighted reward $\alpha \cdot \mathrm{perf} - \beta \cdot \mathrm{cost}$. We sweep five weight settings from the quality-only $(\alpha, \beta) = (1.0, 0.0)$ to the heavily cost-weighted $(0.2, 0.8)$. Multi-round and RL-based routers cannot optimize this weighted objective and are therefore run once under a single configuration. On the personalized track, answers are scored by a persona-conditioned LLM judge (DeepSeek-V3.1) as win, tie, or loss ($1$, $0.5$, $0$), and the judge's own cost is excluded from the reported cost. 

\subsection{Main Results}
\label{sec:results}

\begin{table}[t]
\centering
\caption{\textbf{Results on xRouteBench under the performance-first setting $(\alpha, \beta) = (1.0, 0.0)$.} Scores are reported across the Generic LLM Tasks, memory, vision, and time-series
tracks, together with their average. Following the original implementations where applicable, all multi-turn routers use Qwen2.5-3B-Instruct as the base model. Top two results are highlighted in \textbf{bold} and \underline{underline}.}
\label{tab:main}
\small
\resizebox{\linewidth}{!}{%
\begin{tabular}{lcccccccc}
\toprule
\multirow{2}{*}{\textbf{Router}} & \multirow{2}{*}{\textbf{Generic LLM Tasks}} & \multicolumn{2}{c}{\textbf{Memory}} & \multicolumn{3}{c}{\textbf{Vision}} & \multirow{2}{*}{\textbf{TimeSeries}} & \multirow{2}{*}{\textbf{Avg}} \\
\cmidrule(lr){3-4}\cmidrule(lr){5-7}
 & & \textbf{LoCoMo} & \textbf{LongMemEval} & \textbf{Geometry3K} & \textbf{MathVista} & \textbf{Video} & & \\
\midrule
\rowcolor{barGray} \multicolumn{9}{c}{\textit{Rule-based baselines}} \\
Smallest-LLM & 57.55 & 25.44 & 36.77 & 27.87 & 35.00 & \textbf{33.33} & 49.61 & 37.94 \\
Largest-LLM & 70.29 & 26.59 & 35.57 & 37.70 & 33.00 & 22.22 & 45.67 & 38.72 \\
\midrule
\rowcolor{barGray} \multicolumn{9}{c}{\textit{Single-turn routers}} \\
$k$NNRouter & 71.37 & 25.24 & \textbf{38.74} & 31.15 & 41.00 & \underline{29.63} & 51.97 & 41.30 \\
SVMRouter & 74.21 & \textbf{27.64} & \underline{38.68} & \underline{42.62} & \underline{47.00} & \underline{29.63} & 55.91 & \underline{45.10} \\
MLPRouter & 68.12 & \underline{26.78} & 32.27 & 27.87 & 34.00 & \underline{29.63} & 56.69 & 39.34 \\
MFRouter & 67.23 & 24.49 & 34.91 & 40.98 & 29.00 & 22.22 & 51.97 & 38.69 \\
EloRouter & 64.15 & 25.70 & 37.27 & \textbf{45.90} & \textbf{50.00} & 25.93 & \textbf{63.78} & 44.68 \\
Hybrid LLM & 64.68 & 25.89 & 36.56 & 32.79 & 37.00 & \textbf{33.33} & 51.18 & 40.20 \\
RouterDC & \textbf{80.56} & 24.93 & 36.77 & 16.39 & 24.00 & 25.93 & 45.67 & 36.32 \\
GraphRouter & \underline{80.54} & 25.94 & 33.93 & \underline{42.62} & \textbf{50.00} & 22.22 & \underline{62.99} & \textbf{45.46} \\
CausalLM & 66.90 & 25.40 & 37.60 & 24.60 & 34.00 & \textbf{33.33} & 45.70 & 38.22 \\
\midrule
\rowcolor{barGray} \multicolumn{9}{c}{\textit{Multi-turn routers}} \\
Router-R1 & 35.64 & 24.60 & 17.28 & 14.75 & 18.00 & 22.22 & 23.62 & 22.30 \\
$k$NN-MultiRound & 13.99 & 24.70 & 18.32 & 16.39 & 30.00 & 25.93 & 33.07 & 23.20 \\
LLM-MultiRound & 12.98 & 24.60 & 17.44 & 14.29 & 31.03 & 25.93 & 30.33 & 22.37 \\
\bottomrule
\end{tabular}%
}
\vspace{-10pt}
\end{table}

Table~\ref{tab:main} and Table~\ref{tab:personalized} report the results under the performance-only setting. Based on these results, we have the following key observations:

\noindent\textbf{No single router dominates across all tasks}: The winner router varies across tasks. For example, RouterDC performs best on Generic LLM Tasks and SVMRouter on LoCoMo. Though GraphRouter attains the best average on xRouteBench, it did not consistently outperform other routers in all tasks.

\noindent\textbf{Multi-turn routing does not consistently outperform single-turn routing}:
Across all benchmarks, multiple rounds of routing and aggregation provide no consistent gain over a single routing decision. For many queries, one well-chosen route is sufficient, whereas additional rounds introduce redundant information and computational overhead. Multi-turn routers also rely on a base model (Qwen2.5-3B-Instruct) to decompose queries and aggregate responses, making their performance sensitive to the capabilities of this model. These results highlight the need for better sufficiency estimation, early stopping, and more effective decomposition and aggregation.

\begin{wraptable}{r}{0.52\textwidth}
\vspace{-12pt}
\centering
\caption{\textbf{Performance comparison on the personalized track.} Top two results are highlighted in \textbf{bold} and \underline{underline}.}
\label{tab:personalized}
\small
\setlength{\tabcolsep}{4pt}
\begin{tabular}{lc|lc}
\toprule
\textbf{Router} & \textbf{Acc.} & \textbf{Router} & \textbf{Acc.} \\
\midrule
GMTRouter & \textbf{68.78} & RouterDC & 56.44 \\
PersonalizedRouter & \underline{67.86} & MFRouter & 54.39 \\
EloRouter & 66.40 & MLPRouter & 52.93 \\
GraphRouter & 65.23 & $k$NNRouter & 51.76 \\
SVMRouter & 65.08 & CausalLM & 46.78 \\
Largest-LLM & 58.05 & Router-R1 & 45.46 \\
Hybrid LLM & 57.91 & Smallest-LLM & 42.53 \\
\bottomrule
\end{tabular}
\vspace{-10pt}
\end{wraptable}

\noindent\textbf{Conditioning on user context helps, but how it is modeled matters}:
Table~\ref{tab:personalized} reports persona-judge accuracy on the personalized track. GMTRouter achieves the highest accuracy of 68.78, outperforming PersonalizedRouter (67.86) and the best user-agnostic router, EloRouter (66.40). The strong performance of both personalized methods confirms the benefit of conditioning routing decisions on user context, while GMTRouter's additional 0.92-point gain over PersonalizedRouter shows that the way user context is encoded and integrated remains important.

\subsection{Performance--Cost Trade-offs}
\label{sec:tradeoff}

\begin{figure}[t]
\centering
\includegraphics[width=\linewidth]{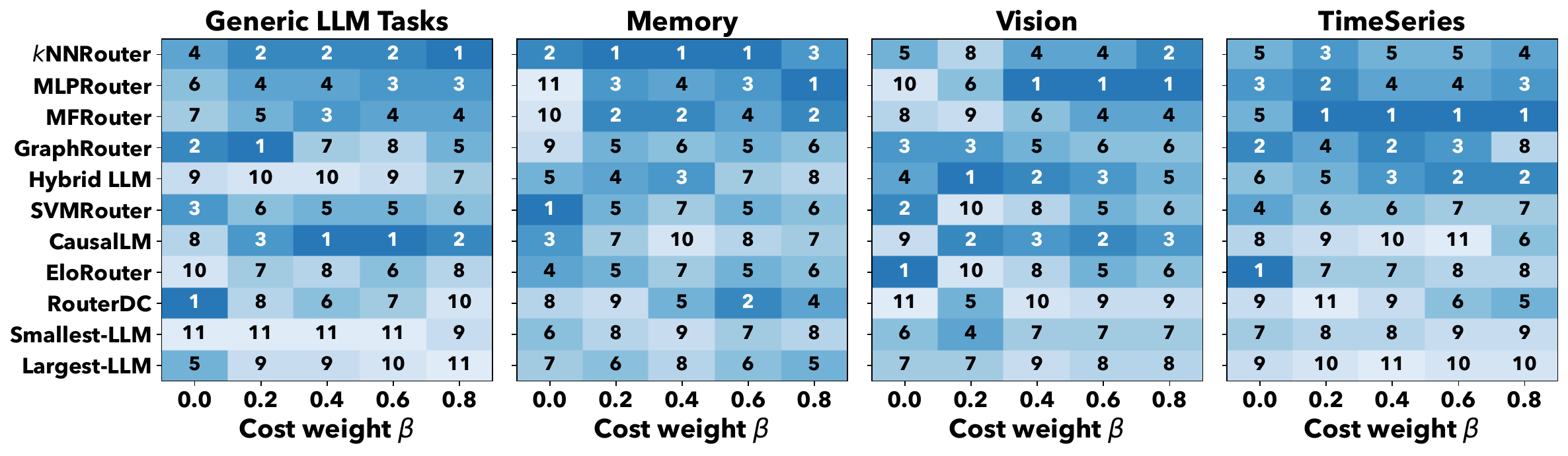}
\caption{\textbf{Router rankings across the Generic LLM Tasks, memory, vision, and time-series tracks as the cost weight $\beta$ increases.} Each cell gives a router's rank under the weighted performance--cost objective, with smaller rank values indicating better performance.}
\label{fig:rank-beta}
\vspace{-10pt}
\end{figure}

\noindent\textbf{No single router is best under every performance--cost trade-off.} Figure~\ref{fig:rank-beta} ranks the routers by reward within each category as the cost weight $\beta$ grows, and the rankings shift dramatically along the sweep. RouterDC tops Generic LLM Tasks when only quality matters, yet falls to tenth of eleven under the most cost-sensitive setting; EloRouter leads Vision and TimeSeries at $\beta = 0$ but drops out of the lead once cost enters the objective. However, weakness at one operating point does not imply weakness at another, as MLPRouter sits near the bottom of Vision under the quality-first setting yet becomes the best choice there for every $\beta \geq 0.4$. Therefore, it is practical to choose the router that matches the performance--cost requirements of the deployment at hand.

\begin{wrapfigure}{r}{0.5\textwidth}
\centering
\vspace{-15pt}
\includegraphics[width=1\linewidth]{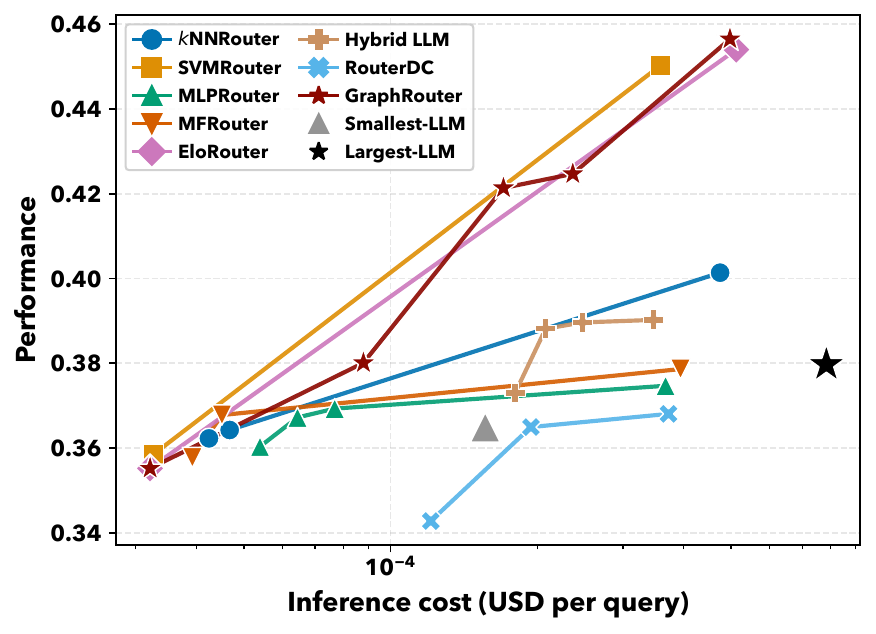}
\vspace{-10pt}
\caption{\textbf{Performance--cost trade-offs of routers averaged across the xRouteBench tracks.} Each point represents an operating setting with a different cost weight $\beta$, where higher performance and lower per-query inference cost are preferred.}
\label{fig:perf-price-all}
\vspace{-30pt}
\end{wrapfigure}

\noindent\textbf{Increasing inference cost leads to improved performance.} Figure~\ref{fig:perf-price-all} presents the trade-off between performance and inference cost. For most routers, performance and cost exhibit a clear positive correlation, with the operating points rising from the low-cost to the high-cost end. This is because increasing the inference budget unlocks more powerful and expensive models, which perform better. Meanwhile, always calling the largest model incurs the highest cost yet delivers only mediocre performance and is dominated by the learned routes, since many queries that the largest model fails are solved by smaller and cheaper ones. This confirms that no single model covers all queries, which is exactly the headroom that routing exploits.

\subsection{Routing in Deployment: Real Users and Multi-Agent Systems}
\label{sec:extension-study}
\begin{wraptable}{r}{0.52\textwidth}
\vspace{-20pt}
\centering
\caption{\textbf{Router performance on held-out real-user sessions collected through the Slack deployment.} Accuracy measures how often each router's model selection agrees with the users' pairwise preferences.}
\label{tab:personalized_human}
\small
\setlength{\tabcolsep}{4pt}
\begin{tabular}{lc|lc}
\toprule
\textbf{Router} & \textbf{Acc.} & \textbf{Router} & \textbf{Acc.} \\
\midrule
PersonalizedRouter & \textbf{83.05} & RouterDC & 65.25 \\
EloRouter & \underline{82.20} & $k$NNRouter & 60.17 \\
MLPRouter & 78.81 & $k$NN-MultiRound & 60.17 \\
SVMRouter & 77.12 & Smallest-LLM & 55.08 \\
Hybrid LLM & 73.73 & MFRouter & 51.69 \\
GMTRouter & 70.70 & Largest-LLM & 41.53 \\
GraphRouter & 67.17 & CausalLM & 27.97 \\
\bottomrule
\end{tabular}
\vspace{-20pt}
\end{wraptable}

The deployment layer of \method\ carries routers beyond static benchmarks. We study two settings it enables: routing for real users served through OpenClaw and routing inside multi-agent systems.

\noindent\textbf{Routing for real users.}
Using the OpenClaw server of \method, we deploy the routing stack behind Slack and collect live preference feedback. 15 users contribute 40 sessions of 1 to 12 turns, totaling 234 pairwise records. For each query, two models are sampled from a pool of ten candidates, their answers are shown in randomized positions, and the user marks one as better or declares a tie. We split by session into 32 training sessions and 8 test sessions, train every router on the human training split, and score how often its selection matches the human preference on held-out sessions. Table~\ref{tab:personalized_human} shows that PersonalizedRouter leads at 83.05, while the fine-tuned CausalLM router ranks last. The simulated ranking does not fully transfer, as GMTRouter, the winner under the persona judge, drops to sixth on real users, showing that it matters to validate personalized routers against real feedback.

\noindent\textbf{Routing inside multi-agent systems.}
A multi-agent system (MAS) is conventionally instantiated with a single base model shared by every agent. We instead treat model choice as a per-agent decision, where a router receives the prompt of each functional node and selects the most suitable LLM for that call. Since node prompts differ substantially, covering planning, execution, and verification instructions, this setting stress-tests how well a router trained on ordinary queries generalizes. Following MultiAgentBench \citep{zhu2025multiagentbench} and GraphPlanner \citep{feng2026graphplanner}, we instantiate five coordination topologies, namely Star, Tree, Graph, Chain, and Plan-Exec-Sum, illustrated in Figure~\ref{fig:multiagent} with node roles detailed in Appendix~\ref{app:mas}. The learning-based routers are trained on the Generic LLM Tasks training split, each test query is then fed through the full MAS, and the final MAS answer is scored by the task metric. Table~\ref{tab:mas} shows that routing every node pays off, as six of the seven learned routers beat always selecting the largest model on average, with MFRouter attaining the best average of 76.48 against 71.48.

\begin{figure}[t]
\centering
\includegraphics[width=\linewidth]{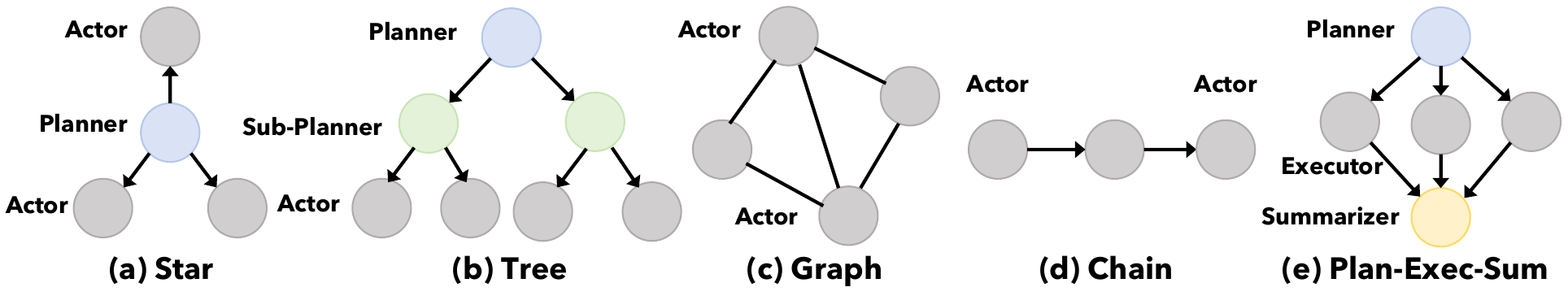}
\caption{\textbf{Representative multi-agent system architectures and coordination topologies:} (a) star-based centralized coordination, (b) hierarchical tree-based delegation, (c) graph-based peer interaction, (d) sequential chain collaboration, and (e) planner–executor–summarizer workflow.}
\label{fig:multiagent}
\end{figure}

\begin{table}[t]
\centering
\caption{\textbf{Router performance on the Generic LLM Tasks test split when each node in a multi-agent system is routed independently.} Results are reported across five coordination topologies, with the final column showing the average performance.}
\label{tab:mas}
\small
\setlength{\tabcolsep}{6pt}
\begin{tabular}{lcccccc}
\toprule
\textbf{Router} & \textbf{Star} & \textbf{Tree} & \textbf{Graph} & \textbf{Chain} & \textbf{Plan-Exec-Sum} & \textbf{Avg} \\
\midrule
Largest-LLM & 69.00 & 67.00 & 77.20 & 69.00 & \underline{75.20} & 71.48 \\
\midrule
$k$NNRouter & 74.80 & \underline{78.60} & 78.60 & 76.60 & 71.80 & 76.08 \\
SVMRouter & \underline{76.20} & 75.60 & \underline{80.00} & 74.40 & \underline{75.20} & \underline{76.28} \\
MLPRouter & 75.40 & 76.60 & 76.80 & \underline{78.00} & 71.40 & 75.64 \\
MFRouter & 75.40 & 74.20 & \textbf{81.00} & \textbf{78.60} & 73.20 & \textbf{76.48} \\
EloRouter & 73.80 & 72.40 & 78.60 & 76.60 & \underline{75.20} & 75.32 \\
GraphRouter & 68.20 & 70.80 & 66.20 & 72.00 & 69.00 & 69.24 \\
RouterDC & \textbf{77.60} & \textbf{79.60} & 74.20 & 72.00 & \textbf{76.20} & 75.92 \\
\bottomrule
\end{tabular}
\vspace{-10pt}
\end{table}
\section{Related Work}
\label{sec:related}

\noindent\textbf{LLM Routing.}
Prior routers can be read along the axes of our formulation. Single-turn routers differ mainly in how they encode a query and score candidates, from quality predictors that arbitrate between a weak and a strong model \citep{ding2024hybrid, ong2024routellm} and classifiers over frozen embeddings \citep{shnitzer2023large, stripelis2024tensoropera, li2025rethinking}, to reward-guided rankings \citep{lu2024routing}, prompt-conditioned preference models \citep{frick2025prompt}, contrastive query--model matching \citep{chen2024routerdc}, learned model embeddings \citep{zhuang2025embedllm}, graph-based scorers \citep{feng2024graphrouter}, and online methods under bandit feedback \citep{dai2024cost, wang2025mixllm}. Multi-turn routers instead enrich the state across rounds, whether as cascades that escalate upon failed verification \citep{chen2023frugalgpt, aggarwal2024automix} or as agentic routers that decompose a query and route sub-queries with reinforcement learning \citep{zhang2025router, feng2026graphplanner}. Personalized routers add the user to the state and learn from preference feedback \citep{xie2025gmtrouter, dai2025personalizedrouter, yu2026tsrouter}. Each is developed in its own formalism and evaluated on its own stack; \method\ instead expresses them as instantiations of a single sequential decision process behind one interface, so single-turn, multi-turn, and personalized routers meet on the same performance--cost frontier.

\noindent\textbf{Routing Benchmarks and Evaluation.}
RouterBench \citep{hu2024routerbench} precomputes candidate responses over a fixed pool, RouterEval \citep{huang2025routereval} aggregates large-scale performance records for routing study, preference data from Chatbot Arena has served as routing supervision \citep{ong2024routellm}, and recent benchmarks extend routing to vision--language pools \citep{huang2025vl, ma2026mmr}. However, existing benchmarks each target a single scenario, one-shot text or image QA, with fixed pools, and provide no pipeline for constructing supervision on new tasks or candidate sets. \method\ closes this gap with automatic supervision construction and cost-aware evaluation over configurable pools, and xRouteBench spans generic LLM tasks, memory-augmented, vision (image and video), time-series, and personalized scenarios under one protocol.

\section{Conclusion}
\label{sec:conclusion}

We introduced \method, a unified framework for LLM routing that casts single-turn, multi-turn, and personalized routing as instances of a common sequential decision process. \method also provides an automatic pipeline for constructing routing supervision and evaluation for new tasks and candidate pools, the multi-scenario xRouteBench benchmark, and an open-source library that implements more than 16 routers behind a unified interface and supports deployment to real users and multi-agent systems. We hope \method will serve as a common foundation for developing, evaluating, and deploying LLM routers.



\bibliography{iclr2026_conference}
\bibliographystyle{iclr2026_conference}
\appendix
\newpage

\begin{center}
    {\bf\Large Appendix}
\end{center}

\startcontents[sections]
\printcontents[sections]{l}{1}{\setcounter{tocdepth}{3}}

\newpage

\section{Benchmark Details} \label{app:benchmarks}

Table~\ref{tab:app-benchmarks} lists the eight test sets of xRouteBench, their categories, sizes, and metrics, totaling 4{,}767 test queries. The Generic LLM Tasks track further decomposes into 13 subtasks, and the memory datasets are retrieved with top-5 RAG over turn pairs.

\begin{table}[h]
\centering
\caption{\textbf{The eight test sets of xRouteBench.} Sizes are the number of test queries; metrics are exact match (EM), multiple-choice accuracy (MC), token-level F1, execution-based code pass rate, math answer matching, and a persona-conditioned LLM judge.}
\label{tab:app-benchmarks}
\small
\resizebox{\linewidth}{!}{%
\begin{tabular}{lllrl}
\toprule
\textbf{Category} & \textbf{Test set} & \textbf{Content} & \textbf{\#Test} & \textbf{Metric} \\
\midrule
Generic LLM Tasks & Generic mix & 13 subtasks & 3{,}729 & EM/MC/F1/GSM8K/MATH/code \\
\midrule
\multirow{2}{*}{Memory} & LoCoMo & long-conversation QA & 314 & F1 \\
 & LongMemEval & long-term memory QA & 101 & F1 \\
\midrule
TimeSeries & TimeSeries & 7 reasoning skills & 127 & MC \\
\midrule
\multirow{3}{*}{Vision} & Geometry3K & geometry math (image) & 61 & EM \\
 & MathVista & visual math reasoning & 100 & EM/MC \\
 & Charades-Ego & egocentric video & 27 & EM \\
\midrule
Personalized & Chatbot Arena / MT-Bench & preference prompts & 308 & LLM judge \\
\midrule
\textbf{Total} & & & \textbf{4{,}767} & \\
\bottomrule
\end{tabular}
}
\end{table}

Table~\ref{tab:app-classic} details the 13 subtasks that make up the Generic LLM Tasks track.

\begin{table}[h]
\centering
\caption{\textbf{Composition of the Generic LLM Tasks track.} The table lists the 13 subtasks, their target skills, and the number of test queries, totaling 3{,}729 examples.}
\label{tab:app-classic}
\small
\setlength{\tabcolsep}{6pt}
\begin{tabular}{llr}
\toprule
\textbf{Subtask} & \textbf{Skill} & \textbf{\#Test} \\
\midrule
MBPP & code generation & 500 \\
MATH & mathematical reasoning & 500 \\
GSM8K & mathematical reasoning & 500 \\
MMLU-Pro & knowledge QA & 500 \\
OpenBookQA & knowledge QA & 500 \\
ARC-Challenge & knowledge QA & 500 \\
MMLU & knowledge QA & 500 \\
CommonsenseQA & commonsense QA & 50 \\
BoolQ & commonsense QA & 50 \\
SQuAD & reading comprehension & 50 \\
HellaSwag & commonsense QA & 50 \\
HumanEval & code generation & 16 \\
AIME (2020--2024) & competition math & 13 \\
\bottomrule
\end{tabular}
\end{table}

\subsection{Generic LLM Tasks} \label{app:general_nlp}

The Generic LLM Tasks track is deliberately a mixture rather than a single task family. It places knowledge questions, commonsense inference, reading comprehension, mathematical reasoning, and code generation behind the same routing interface. All examples use text input, but an ARC-Challenge question, an AIME problem, and a Python synthesis task reward very different capabilities and impose different output constraints. The track therefore asks whether a router can recognize these distinctions within an apparently uniform text setting, rather than defaulting to one model for every natural language request. Table~\ref{tab:app-classic} lists the 13 source benchmarks and their test sizes.

We serialize each example as one fixed text query before collecting candidate responses. A system instruction specific to the task states the required response format, and the user problem follows it. Choice items include their options. Mathematics items request a final answer in the expected form. Code generation items provide the programming task and tests, while SQuAD supplies its passage and question. We score each response with the source benchmark's native objective. We use choice accuracy for knowledge and commonsense tasks, token-level F1 for SQuAD, answer matching for GSM8K, MATH, and AIME, and execution-based pass rates for MBPP and HumanEval.

\subsection{Long Context Conversational Memory} \label{app:memory}

The Memory track fixes retrieval and asks a narrower routing question. Given the same retrieved conversational evidence, which model is most likely to use it correctly? LoCoMo and LongMemEval contain questions whose supporting facts may be separated from the query by many turns or sessions. Some can be answered by recovering one explicit fact, whereas others require combining facts across sessions or resolving a later update against an earlier statement. This distinction is useful for routing because it separates the cost of carrying a long history from the ability to interpret the evidence selected from it.

We construct every memory query through the same retrieval pipeline. Each conversation history is divided into adjacent turn pairs that retain speaker and date information. A fixed Contriever \citep{izacard2022unsupervised} encoder embeds the question and the turn pairs, and the five most similar pairs are inserted into a brief answer prompt. LongMemEval additionally includes the date of the question, which provides the temporal reference needed for its time-sensitive items. Every candidate therefore receives the same retrieved evidence for a query. Differences in score reflect its use of that evidence rather than a different retrieval result. We evaluate both datasets with token-level F1.

\subsection{Time Series Pattern Reasoning} \label{app:timeseries}

The TimeSeries track is built from TSRBench and covers anomaly detection, similarity analysis, noise understanding, pattern recognition, inductive reasoning, causality analysis, and event prediction. These problems are numerical, but they often require recognizing structure that is easier to see as a shape than as a list of values. A local spike may signal an anomaly, while periodicity, changes in trend, or agreement between two series emerge over a longer range. The track asks whether a router can distinguish the models that handle these different forms of temporal reasoning well.

We convert every series into a common text query before collecting candidate responses. First, we render each series as a line chart and pass the chart to a fixed Gemma-3-27B-IT \citep{team2025gemma} captioner, which produces a paragraph description of its visible temporal pattern. We append these descriptions and the raw numerical values to the original question and its answer options. The raw values preserve exact numerical evidence and are truncated after 200 values when necessary, while the descriptions make higher-level structure explicit. The charts are used only in this offline captioning step. Every candidate model receives the same text query rather than an image. Models return an option letter, and we score the responses with multiple-choice accuracy.

\subsection{Visual Mathematical Reasoning}  \label{app:math_image}

The Visual Mathematical Reasoning track routes image-grounded mathematics drawn from Geometry3K \citep{lu2021inter} and MathVista \citep{lu2024mathvista}, where a geometry diagram or a scientific figure carries information the question depends on. The strongest candidate on these problems can differ from the strongest on text mathematics, so the track probes whether a router follows per-model strength as the query type changes. We render each problem to text before it reaches the pool, describing its image with a fixed vision-language model and appending the description to the question, so every candidate reasons over one identical query. Rendering the image once holds perception constant across the pool and turns visual mathematics into a query the whole pool can answer, so the track scores the reasoning quality of each candidate on the same input. These rendered queries read as machine-written descriptions of a figure, a distribution that departs from the natural questions the routers are trained on, so the track also tests how well a router generalizes when the query type shifts.

\begin{figure}[H]
\centering
\begin{minipage}[c]{0.24\linewidth}
\centering
\includegraphics[width=0.9\linewidth]{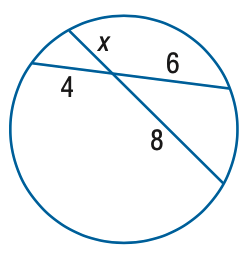}\\[3pt]
{\footnotesize (a) Geometry3K, which asks for $x$.}
\end{minipage}
\hfill
\begin{minipage}[c]{0.72\linewidth}
\begin{mybox}[Gemma-3-27B-IT description]
\footnotesize The diagram shows a circle with two intersecting chords. One chord is divided into segments of length $4$ and $6$, and the other into segments of length $x$ and $8$. The label $x$ marks one of these segments, and the two chords cross at a point inside the circle.
\end{mybox}
\end{minipage}

\vspace{6pt}

\begin{minipage}[c]{0.40\linewidth}
\centering
\includegraphics[width=\linewidth]{figs/mathvista.png}\\[3pt]
{\footnotesize (b) MathVista, which asks for the spring compression $d$.}
\end{minipage}
\hfill
\begin{minipage}[c]{0.56\linewidth}
\begin{mybox}[Gemma-3-27B-IT description]
\footnotesize A block of mass $m$ slides to the left along a horizontal frictionless surface with velocity $v$ and approaches a coiled spring of constant $k$ fixed to a wall. Labels mark where the block first touches the spring and where it stops, and $d$ denotes the distance between these two points. A caption states that the spring force does negative work, decreasing speed and kinetic energy.
\end{mybox}
\end{minipage}
\caption{\textbf{One example from each dataset in the Visual Reasoning track, shown with the description that Gemma-3-27B-IT produces for its image.} Geometry3K (a) provides a geometry diagram, and MathVista (b) provides a scientific figure. Each description is appended to the problem text to form the query the router sees.}
\label{fig:multimodal_examples}
\end{figure}

A single frozen vision-language model, Gemma-3-27B-IT by default, produces every description, so all candidates see the same rendering of a given image. Its prompt asks it to report every visible number, symbol, angle, length, and relationship and to withhold the solution, which keeps the answer out of the query and leaves the reasoning to the routed model. The returned text is appended to the original problem, from which Geometry3K's image placeholder is removed, to form the routed query. Geometry3K is graded by math answer matching against its numeric answer, and MathVista is graded by exact match on its open-ended items and by multiple-choice accuracy on its choice items, following the question-type field of each query. Every query keeps its ground-truth answer and, where present, its answer choices, so the scored responses populate the same query--model matrix of quality and cost that supervises every other track (\S\ref{sec:protocol}). Figure~\ref{fig:multimodal_examples} shows one example from each dataset with the description Gemma-3-27B-IT produces for it.

\subsection{Multi-View Video Recognition} \label{app:video}

\begin{figure}[H]
\centering
\setlength{\tabcolsep}{2pt}
\renewcommand{\arraystretch}{0.7}
\begin{tabular}{c ccccc}
 & \multicolumn{5}{c}{\footnotesize time $\rightarrow$} \\
\rotatebox{90}{\footnotesize First-person} & \includegraphics[width=0.18\linewidth]{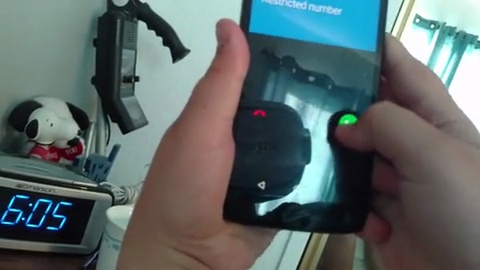} & \includegraphics[width=0.18\linewidth]{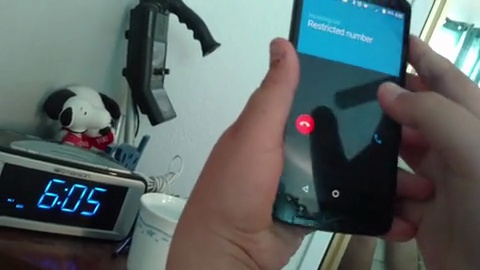} & \includegraphics[width=0.18\linewidth]{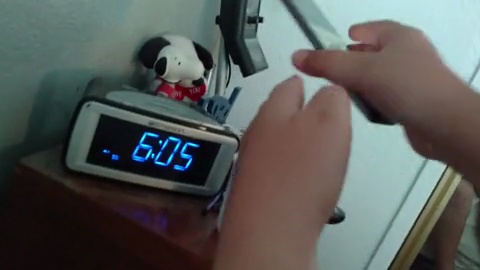} & \includegraphics[width=0.18\linewidth]{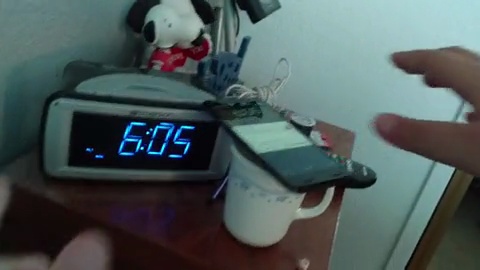} & \includegraphics[width=0.18\linewidth]{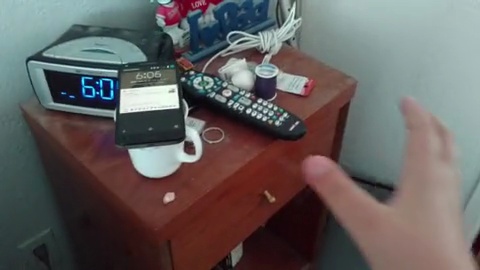} \\
\rotatebox{90}{\footnotesize Third-person} & \includegraphics[width=0.18\linewidth]{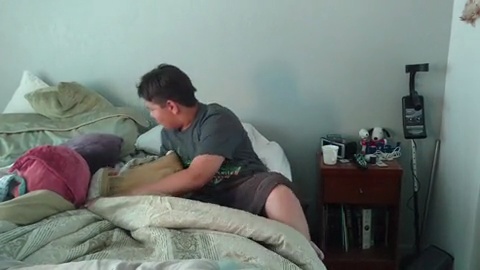} & \includegraphics[width=0.18\linewidth]{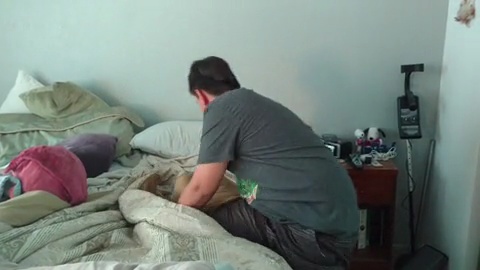} & \includegraphics[width=0.18\linewidth]{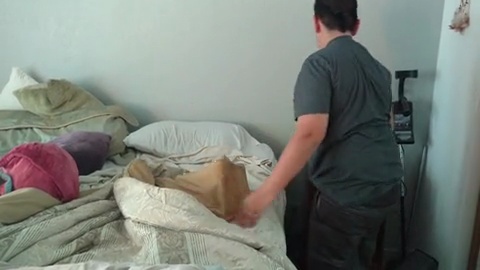} & \includegraphics[width=0.18\linewidth]{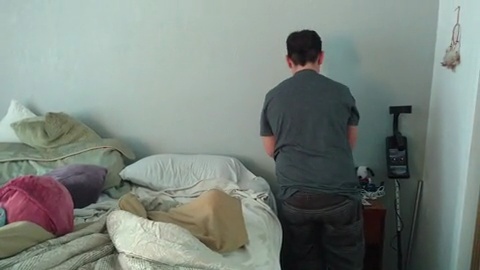} & \includegraphics[width=0.18\linewidth]{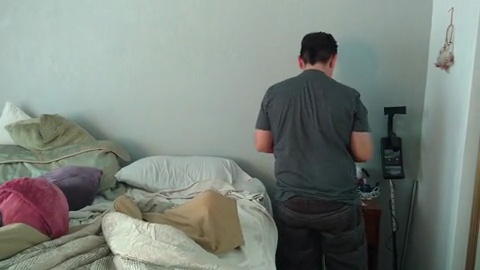} \\
\end{tabular}

\vspace{4pt}
\begin{mybox}[Gemma-3-27B-IT description]
\footnotesize \textit{Placeholder for the structured description the vision-language model produces from the available views.}
\end{mybox}
\caption{\textbf{A multi-view sample from Charades-Ego \citep{sigurdsson2018charades}, recorded from a first-person egocentric camera (top) and a third-person exocentric camera (bottom).} Each row is a viewpoint and each column is a sampled time step. Our transformation samples these frames, describes the available views with a vision-language model, and routes the resulting text query, randomly withholding one view to form single-view and dual-view queries.}
\label{fig:charades_sample}
\end{figure}

The video track builds on Charades-Ego \citep{sigurdsson2018charades}, which records each activity simultaneously from a first-person egocentric camera worn by the actor and a third-person exocentric camera facing the scene (Figure~\ref{fig:charades_sample}). The two viewpoints carry different evidence, and a deployment often captures only one of them, so the routing decision here depends on which views a query provides. We describe every available view in text and route the merged description, so the whole candidate pool answers the same query whether one view or both are present. This makes the video track the regime in xRouteBench where the available modality varies from query to query, and it tests whether a router still selects the right model when the visual evidence is partial.

We build three classification tasks from the annotations, predicting the activity, the verb, or the object of the depicted action, each answered by a compact identifier drawn from the task label inventory, with the verb and object labels recovered from the Charades action mapping. For each paired clip we match the two views by their shared identifier and select the action segment whose egocentric and exocentric occurrences overlap most closely in time, which keeps both views on the same moment, and we then keep the first-person view, the third-person view, or both at random to populate the single-view and dual-view regimes. From each retained view we sample frames inside the aligned window and pass them to the vision-language model, which returns a structured description of the actor's motion, the handled objects, and the scene. These descriptions are merged into one query that states the task, lists the candidate identifiers with their names, and requests an identifier as the answer, holding every candidate to the same text input.

\subsection{Personalized Dialogue Preference} \label{app:personalized}

The Personalized track combines open-ended dialogue from MT-Bench and Chatbot Arena. Each query retains its original message sequence, including preceding system, user, and assistant messages. MT-Bench contributes instruction following conversations in which a later turn depends on an earlier exchange, while Chatbot Arena contributes natural user requests across diverse conversational settings. Rather than assuming that one answer is universally better, the track constructs supervision from the preference that a particular user profile would express after seeing two answers.

The user profiles used for preference elicitation are drawn from PersonaHub \citep{tencent2024personahub}. We sample 200 personas from this resource for data collection. Figure~\ref{fig:personahub_personas} presents ten examples from the full persona set used during data collection.

\begin{figure}[H]
\centering
\begin{minipage}{0.98\linewidth}
\begin{mybox}[Ten example personas used to condition the judge]
\footnotesize
\begin{tabular}{@{}p{0.47\linewidth}@{\hspace{0.03\linewidth}}p{0.47\linewidth}@{}}
\textbf{1.} A 71-year-old retired nurse from Italy, volunteering in hospice care and advocating for compassionate end-of-life support. & \textbf{6.} A 34-year-old scientist from London who is a social media influencer. \\
\textbf{2.} A 54-year-old divorced mother from Spain, running a successful winery and promoting sustainable viticulture practices. & \textbf{7.} A 41-year-old scientist from London who loves hiking. \\
\textbf{3.} A 63-year-old retired teacher from China, teaching calligraphy and preserving the art form for future generations. & \textbf{8.} An 87-year-old World War II veteran from Poland, sharing stories of his experiences and advocating for peace. \\
\textbf{4.} A 68-year-old retired engineer from Japan, practicing ikebana and teaching the art to younger generations. & \textbf{9.} A 31-year-old social worker from Colombia, supporting victims of domestic violence and fighting for gender equality. \\
\textbf{5.} A 21-year-old photographer from Paris who spends weekends volunteering. & \textbf{10.} A 23-year-old aspiring musician from Brazil, fusing traditional and modern sounds and promoting cultural exchange through music. \\
\end{tabular}
\end{mybox}
\end{minipage}
\caption{\textbf{Ten examples from the 200 PersonaHub personas sampled for preference collection.} For each comparison, DeepSeek-V3.1 receives one selected persona as its role specification when judging the two candidate answers. The candidate models do not receive the persona.}
\label{fig:personahub_personas}
\end{figure}

For each dialogue, we sample two models at random from the candidate pool of 18 models and generate one response from each under the same conversation history. DeepSeek-V3.1 is conditioned on the selected persona to act as the judge, comparing the two answers and returning a preference for either response or a tie. The persona is supplied to the judge only, not to either candidate model. Each comparison is converted into a pairwise supervision record whose label is the final judged user preference. The router learns from these preference outcomes rather than from the persona description itself. This construction keeps the candidate responses comparable while making the routing target sensitive to the user utility represented by the judge.

\section{Router Details} \label{app:routers}

The routers in \method share the candidate pool and task interface described above, but they expose different information to the routing decision. Rule-based baselines select from a fixed property of the pool, while single-turn methods make one model selection from the query and logged routing outcomes. Multi-turn methods additionally condition on intermediate calls and may decide whether further routing is useful. Personalized methods add a user and interaction history, so their target is the model a particular user is most likely to prefer rather than one model that is best on average. Table~\ref{tab:app-routers} summarizes the 17 built-in implementations along these differences.

\begin{table}[t]
\centering
\caption{\textbf{Built-in routers in \method, grouped by routing family.} For each method, the table summarizes the information available to the routing decision (State) and the rule used to select or aggregate candidate models (Selection).}
\label{tab:app-routers}
\small
\setlength{\tabcolsep}{4pt}
\renewcommand{\arraystretch}{1.08}
\newcommand{\routerspace}{\addlinespace[2pt]}
\begin{tabular}{
    p{0.20\linewidth}
    p{0.30\linewidth}
    p{0.32\linewidth}
}
\toprule
\textbf{Router} & \textbf{State} & \textbf{Selection} \\
\midrule
\rowcolor{barGray} \multicolumn{3}{c}{\textit{Rule-based baselines}} \\
Smallest-LLM
& candidate parameter counts
& always selects the smallest candidate \\
\routerspace
Largest-LLM
& candidate parameter counts
& always selects the largest candidate \\
\midrule
\rowcolor{barGray} \multicolumn{3}{c}{\textit{Single-turn routers}} \\
kNNRouter
& query embedding and nearby logged queries
& votes over the models preferred by nearest neighbors \\
\routerspace
SVMRouter
& query embedding
& kernel classifier predicts a candidate \\
\routerspace
MLPRouter
& query embedding
& MLP classifier predicts a candidate \\
\routerspace
MFRouter
& query and model latent factors
& ranks candidates by their interaction score \\
\routerspace
EloRouter
& logged pairwise model outcomes
& always selects the highest-rated candidate \\
\routerspace
RouterDC
& query and candidate representations
& contrastive query--model matching score \\
\routerspace
Hybrid LLM
& query embedding and a small/large model pair
& predicts whether the small model is sufficient \\
\routerspace
AutoMix
& small-model draft and verification signal
& accepts the draft or escalates to the large model \\
\routerspace
GraphRouter
& query--model interaction graph
& predicts performance on query--model edges \\
\routerspace
CausalLM Router
& textual query and candidate list
& generates the selected model name \\
\midrule
\rowcolor{barGray} \multicolumn{3}{c}{\textit{Multi-turn routers}} \\
Router-R1
& query and accumulated search results
& iteratively searches specialists or terminates and aggregates \\
\routerspace
kNN-MultiRound
& sub-queries and their embeddings
& routes each sub-query with kNN and aggregates the answers \\
\routerspace
LLM-MultiRound
& textual query, decomposition, and candidate list
& an LLM chooses routes for sub-queries and aggregates \\
\midrule
\rowcolor{barGray} \multicolumn{3}{c}{\textit{Personalized routers}} \\
GMTRouter
& user, session, query, model, and response interactions
& predicts user-conditioned model preference \\
\routerspace
PersonalizedRouter
& user features, task description, query, and model
& predicts preference for a user--query pair \\
\bottomrule
\end{tabular}
\end{table}

\subsection{Rule-Based Baselines}

Smallest-LLM and Largest-LLM provide fixed reference points for the learned routers. They ignore the query and always select the candidate with the smallest or largest declared parameter count, respectively. These rules make no attempt to identify per-query model strengths, but they expose the two simple deployment policies against which query-aware routing should be compared: consistently favoring the smallest available model or consistently favoring the largest one.

\subsection{Single-Turn Routers}

Single-turn routers terminate after one model selection, so their differences lie in how they represent a query and score candidates. EloRouter is query-independent, but it replaces a fixed parameter-count rule with a global ranking estimated from pairwise outcomes in the routing data. It therefore captures which model is strongest on average while deliberately discarding the variation between individual queries.

kNNRouter, SVMRouter, and MLPRouter instead make the selection query-specific from its embedding. kNNRouter retrieves similar training queries and transfers their observed best-model choices through a vote or distance-weighted vote, requiring no fitted decision function beyond the stored examples \citep{li2025rethinking,shnitzer2023large}. SVMRouter and MLPRouter fit discriminative boundaries over the same embedding space, using a kernel classifier and a multilayer perceptron, respectively. MFRouter takes a different view of the logged query--model matrix: it learns latent representations for both sides and selects the model with the strongest query--model interaction. These methods all learn from per-query outcomes, but differ in whether the candidate is represented by neighboring solved examples, a classifier label, or a learned latent factor.

The remaining single-turn methods enrich this compatibility score in different ways. RouterDC learns query--model matching with dual contrastive objectives over query--model, query--query, and cluster-level relations \citep{chen2024routerdc}. GraphRouter instead passes information over a query--model interaction graph and predicts the quality of an unobserved query--model edge \citep{feng2024graphrouter}. CausalLM Router verbalizes the query and the available candidates, then fine-tunes a causal language model to generate the selected model name \citep{ong2024routellm}. In this case, the query representation, candidate representation, and selection score are combined within the language model rather than implemented as separate embedding modules.

Hybrid LLM and AutoMix are cost-aware two-model cascades. Hybrid LLM learns from the quality gap between the smallest and largest candidates, then uses a thresholded prediction to decide whether the smaller model is sufficient \citep{ding2024hybrid}. AutoMix first obtains a draft from the smaller model and a verification signal for that draft; it retains the draft when the signal is reliable and otherwise escalates the query to the larger model \citep{aggarwal2024automix}. We list both methods with the single-turn family because they return one final model answer through a fixed cascade, rather than repeatedly choosing among arbitrary candidates and aggregating an open-ended history. Their extra calls are nevertheless part of the inference cost.

\subsection{Multi-Turn Routers}

Multi-turn routers treat intermediate responses as part of the routing state. Router-R1 is a pretrained routing agent that reasons over the query and the results gathered so far. At each step it can issue a \texttt{<search>} call to a suitable specialist, incorporate the returned evidence, or terminate and produce an aggregate answer; its policy is trained with trajectory-level reinforcement learning rather than pointwise best-model labels \citep{zhang2025router}. The cost of its reasoning, search, and final aggregation calls is included in the routed trajectory.

The two multi-round baselines use decomposition more explicitly. kNN-MultiRound first breaks a complex query into a small set of sub-queries, applies the kNN routing rule to each one, and combines the resulting answers. LLM-MultiRound uses an LLM to express both the decomposition and the model choice in text, then executes the selected sub-queries and aggregates their outputs. Both methods can assign different candidates to different parts of a problem, but they also introduce decomposition and aggregation calls that a one-shot router does not pay for.

\subsection{Personalized Routers}

Personalized routers replace the single global notion of quality with a user-conditioned preference. GMTRouter represents users, sessions, queries, candidate models, and responses as nodes in a heterogeneous interaction graph, allowing a decision for the current query to draw on earlier interactions from the same user \citep{xie2025gmtrouter}. PersonalizedRouter similarly uses a graph-based scorer, while explicitly incorporating user features and task descriptions alongside the query and candidate representations \citep{dai2025personalizedrouter}. Both methods are trained from comparisons between candidate answers, so a high score means that a model is predicted to be preferred for this user and query, not simply that it has the highest population-average task score.

\section{Candidate Pool and Pricing} \label{app:pricing}

Table~\ref{tab:app-pricing} lists the 18 candidate models and their per-token prices. Prices are in USD per 1M tokens, with input and output priced separately; the blended average is their mean. The input price spans a roughly $25\times$ range, from \$0.05 to \$1.25 per 1M tokens. The 17-model no-cogito pool drops the most expensive model (cogito-v2-1-671b).

\begin{table}[h]
\centering
\caption{\textbf{The 18 candidate LLMs, sorted by blended average price.} Prices in USD per 1M tokens.}
\label{tab:app-pricing}
\small
\setlength{\tabcolsep}{6pt}
\begin{tabular}{rlrrrl}
\toprule
\textbf{\#} & \textbf{Model} & \textbf{Params} & \textbf{Input} & \textbf{Output} & \textbf{Service} \\
\midrule
1 & gemma-2-9b-it & 9B & 0.10 & 0.10 & NVIDIA \\
2 & llama-3-8b-instruct-lite & 8B & 0.10 & 0.10 & Together \\
3 & gpt-oss-20b & 20B & 0.05 & 0.20 & Together \\
4 & rnj-1-instruct & 15B & 0.15 & 0.15 & Together \\
5 & mistral-7b-instruct-v0.3 & 7B & 0.20 & 0.20 & NVIDIA \\
6 & mistral-small-3-24b-instruct & 24B & 0.10 & 0.30 & Together \\
7 & qwen2.5-7b-instruct & 7B & 0.20 & 0.20 & NVIDIA \\
8 & qwen2.5-7b-instruct-turbo & 7B & 0.30 & 0.30 & Together \\
9 & gpt-oss-120b & 120B & 0.15 & 0.60 & Together \\
10 & llama-4-maverick & 402B & 0.27 & 0.85 & Together \\
11 & mixtral-8x7b-instruct-v0.1 & 46.7B & 0.60 & 0.60 & NVIDIA \\
12 & qwen3-next-80b-a3b-instruct & 80B & 0.15 & 1.50 & Together \\
13 & qwen3-coder-next & 200B & 0.50 & 1.20 & Together \\
14 & llama-3.3-70b-instruct-turbo & 70B & 0.88 & 0.88 & Together \\
15 & llama3-70b-instruct & 70B & 0.90 & 0.90 & NVIDIA \\
16 & deepseek-v3.1 & 671B & 0.60 & 1.70 & Together \\
17 & mixtral-8x22b-instruct-v0.1 & 140.6B & 1.20 & 1.20 & NVIDIA \\
18 & cogito-v2-1-671b & 671B & 1.25 & 1.25 & Together \\
\bottomrule
\end{tabular}
\end{table}

\section{Human Preference Collection on Slack} \label{app:slack}

\begin{figure}[h]
\centering
\includegraphics[width=1.0\linewidth]{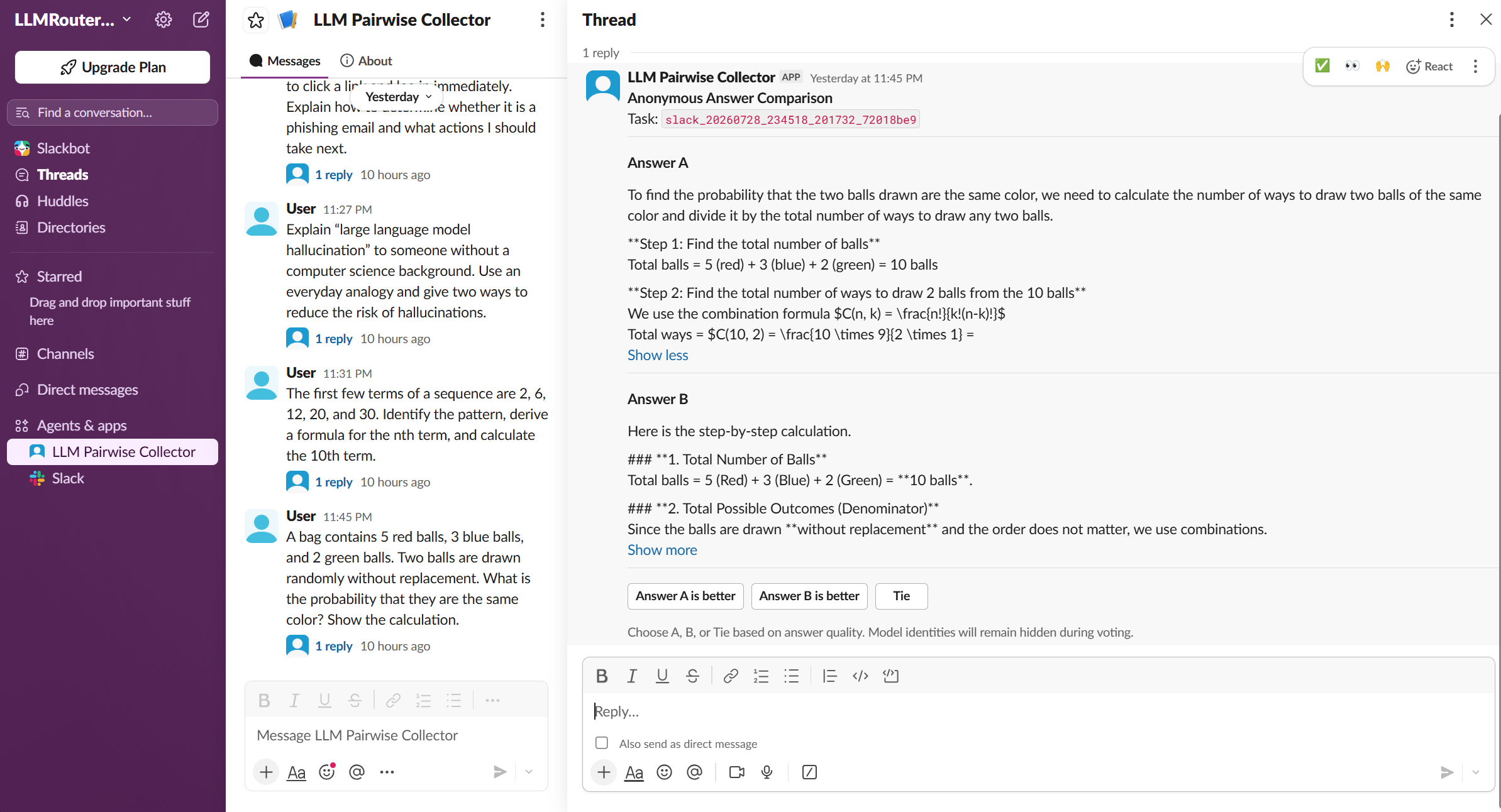}
\caption{\textbf{Slack interface for collecting pairwise user preferences.} Two anonymized model responses are shown as Answer A and Answer B in randomized order, and users select A, B, or a tie for each interaction turn.}
\label{fig:slack}
\end{figure}

The human preference dataset of \S\ref{sec:extension-study} is collected through a Slack application built on the OpenClaw server of \method, as illustrated in Figure~\ref{fig:slack}. A user asks a question directly in Slack, and the system samples two LLMs at random from the ten-candidate pool and generates one answer with each. The conversation view then presents the two responses as anonymized Answer A and Answer B, with their positions randomly shuffled, and the user clicks a button to mark A better, B better, or a tie. In multi-turn sessions the user follows up freely, and every turn is labeled in the same way. Fifteen users contributed 40 sessions and 234 pairwise preference records in total.

\section{Multi-Agent Topologies} \label{app:mas}

Table~\ref{tab:app-mas} summarizes the five coordination topologies used in \S\ref{sec:extension-study}. Star, Tree, Graph, and Chain follow MultiAgentBench \citep{zhu2025multiagentbench}, and Plan-Exec-Sum follows the static router design of GraphPlanner \citep{feng2026graphplanner} with width 3 and depth 1. Star and Plan-Exec-Sum differ in two respects: in Star, the planner produces free-form subtasks and consolidates the actors' outputs itself, whereas in Plan-Exec-Sum the planner decomposes the query into three atomic sub-queries and a dedicated summarizer merges the executors' answers. Every node is replaced by a router that receives the node's prompt and selects a model from the candidate pool for that call. We adopt the topology of GraphPlanner without training its planner, since the training requires live answers from candidate models that have since been retired.

\begin{table}[h]
\centering
\caption{\textbf{The five multi-agent topologies, their structures, and the number of LLM calls per query.}}
\label{tab:app-mas}
\resizebox{\linewidth}{!}{%
\begin{tabular}{llc}
\toprule
\textbf{Topology} & \textbf{Structure} & \textbf{LLM calls} \\
\midrule
Star & planner decomposes $\rightarrow$ 3 actors in parallel $\rightarrow$ planner consolidates & 6 \\
Tree & root planner $\rightarrow$ 2 sub-planners refine $\rightarrow$ 2 actors $\rightarrow$ root consolidates & 7 \\
Graph & 3 actors answer independently $\rightarrow$ one full-communication revision round & 7 \\
Chain & 3 agents relay sequentially, each verifying and improving the previous answer & 4 \\
Plan-Exec-Sum & planner emits 3 atomic sub-queries $\rightarrow$ 3 executors $\rightarrow$ summarizer merges & 6 \\
\bottomrule
\end{tabular}
}
\end{table}

\section{Prompt Usage} \label{app:prompt}

This section reports the fixed prompt templates used to construct and
evaluate xRouteBench, as well as the templates used by routers that make
language-model calls internally. Curly brackets denote instance-dependent
fields. Each prompt is presented in a titled box, combining the fixed
instruction and the instance fields in the order in which the model reads
them, rather than splitting them into separate system and user blocks. In the
implementation, the fixed instruction is sent as a system message when the
API supports that role; otherwise, the two parts are concatenated with
\texttt{[System Instruction]} and \texttt{[User Query]} delimiters. Thus, all
candidate models receive the same logical prompt for a given query.

\newcommand{\prompttext}[1]{{\ttfamily\noindent\detokenize{#1}\par}}
\newcommand{\promptfield}[1]{{\ttfamily\detokenize{{#1}}}}

\subsection{Benchmark-Query Templates}

\paragraph{Generic LLM Tasks.}
The 13 Generic LLM Tasks use seven task-specific templates. The
multiple-choice instruction is shared by MMLU, MMLU-Pro, BoolQ, and
HellaSwag; CommonsenseQA, OpenBookQA, and ARC-Challenge use the same
instruction but retain their source choice formatting.

\begin{mybox}[Standard multiple choice]
\footnotesize
\prompttext{Answer the following multiple-choice question by selecting the correct option
(A, B, C, or D). You MUST put your final answer letter in a parenthesis.}

\medskip

\prompttext{Question:
{question}}

\prompttext{Options:
A. {choice_1}
B. {choice_2}
C. {choice_3}
D. {choice_4}}
\end{mybox}

\begin{mybox}[Source-formatted multiple choice]
\footnotesize
\prompttext{Answer the following multiple-choice question by selecting the correct option
(A, B, C, or D). You MUST put your final answer letter in a parenthesis.}

\medskip

\prompttext{Question: {question}
(A) {choice_1}
(B) {choice_2}
(C) {choice_3}
...}
\end{mybox}

\begin{mybox}[GSM8K]
\footnotesize
\prompttext{Answer the following math question step by step.}
\medskip
\prompttext{Question: {question}}
\end{mybox}

\begin{mybox}[MATH and AIME]
\footnotesize
\prompttext{Answer the following math question. Make sure to put the answer (and only
answer) inside \boxed{}.}
\medskip
\prompttext{Question: {question}}
\end{mybox}

\begin{mybox}[MBPP]
\footnotesize
\prompttext{You are an expert Python programmer. Implement the function with no irrelevant
words or comments. Put your code in this format: [BEGIN] <Your Code> [Done]}
\medskip
\prompttext{Task: {task_description}}

\prompttext{Your code should pass these tests: {tests}}

\end{mybox}

\begin{mybox}[HumanEval]
\footnotesize
\prompttext{You are an expert Python programmer. Implement the function body only. Do not
repeat the function signature or docstring. Put the code in this format:
[BEGIN] <Your Code - function body only> [Done]}
\medskip
\prompttext{Complete the following function: {function_signature_and_docstring}}

\end{mybox}

\begin{mybox}[SQuAD]
\footnotesize
\prompttext{Answer the question using the provided context. If the answer is not in the
context, output noanswer.}

\prompttext{{source query containing the context and question}}
\end{mybox}

\paragraph{Conversational memory.}
For both memory datasets, Contriever retrieves the top five turn-pair chunks
before the following prompt is formed. The retrieval result is fixed across
candidate models for each query. LoCoMo uses:
\begin{mybox}[LoCoMo]
\footnotesize
\prompttext{You answer questions about conversation history. Reply with a short phrase
only.}

\prompttext{Below is relevant information from the conversation history:}

\prompttext{{retrieved turn-pair chunks}}
\medskip
\prompttext{Based on the above context, write an answer in the form of a short phrase for
the following question. Answer with exact words from the context whenever
possible.}

\prompttext{Question: {question} Short answer:}
\end{mybox}
If no chunk is retrieved, the context field is replaced by \texttt{No
relevant information available.} LongMemEval uses the analogous template,
with its question date made explicit:
\begin{mybox}[LongMemEval]
\footnotesize
\prompttext{You answer questions based on chat history. Reply with a short phrase only.}

\prompttext{I will give you several history chats between you and a user. Please answer
the question based on the relevant chat history.}
\medskip
\prompttext{History Chats:}

\prompttext{### Session {index}:
Session Content:
{retrieved chunk}}

\prompttext{...}

\prompttext{Current Date: {current_date}}
\prompttext{Question: {question}}
\prompttext{Short Answer:}
\end{mybox}
Here, an empty history is represented by \texttt{No relevant chat history
available.}

\paragraph{Visual mathematical reasoning.}
Geometry3K and MathVista first use one frozen vision-language model to turn
the image into text. The captioning call is:
\begin{mybox}[Image captioning]
\footnotesize
\prompttext{Describe the image for solving the problem.
Include all visible text, numbers, symbols, angles, lengths, and geometric
relationships.
Be concise and factual. Do not include introductory phrases.}

\prompttext{{image}}
\end{mybox}
The resulting caption is then given to every candidate model with the
following answer prompt:
\begin{mybox}[Visual mathematical reasoning]
\footnotesize
\prompttext{You are an expert in solving math and geometry problems. Think step-by-step,
and when you are ready to provide the final answer, you MUST enclose it in
\boxed{}. }
\medskip
\prompttext{For example, if the answer is 42, write \boxed{42}. If it is a
multiple choice question and the option is C, write \boxed{C}.}
\medskip
\prompttext{Image Description: {caption}}

\prompttext{Answer the following question based on the provided image description:
{question}}
\end{mybox}

\paragraph{Time-series reasoning.}
We first render every series as an image and request a one-paragraph caption.
The captioner receives the following system instruction and one instruction
sampled uniformly from 20 equivalent phrasings (the first phrasing is shown):
\begin{mybox}[Time-series captioning]
\footnotesize
\prompttext{You are a time series captioner.}

\prompttext{Write a paragraph that analyzes the time series.}

\prompttext{{time-series plot}}
\end{mybox}
The other phrasings differ only in wording (e.g., ``Create a detailed
description of the time series in one paragraph'') and impose the same
one-paragraph description requirement. The answer prompt combines the caption
with the raw values:
\begin{mybox}[Time-series reasoning]
\footnotesize
\prompttext{You are a time series analysis expert. Answer the multiple-choice question by
providing the correct option letter (e.g., A, B, C, or D). Be concise.}
\medskip
\prompttext{1. Time Series Description: (1) {series_name}: {caption}, ...}
\prompttext{2. Raw time series: (1) {series_name}: [{values}], ...}
\medskip
\prompttext{based on the one of the following information to reason the final answer:
{original_question}}
\end{mybox}
For perception and causality-analysis items, this two-source block replaces
the \texttt{<ts><ts/>} marker in the source question; it is appended for the
other two task types.

\paragraph{Multi-view video recognition.}
Five frames from each available egocentric or exocentric view are summarized
independently before classification. The video captioner receives:
\begin{mybox}[Video captioning]
\footnotesize
\prompttext{You are given multiple frames sampled from a short time window of a video.}
\medskip
\prompttext{Return ONLY a JSON object with the following keys (no extra text):}
\prompttext{  "motion": string,}
\prompttext{  "objects": [string],}
\prompttext{  "summary": string}
\medskip
\prompttext{Definitions:}
\medskip
\prompttext{motion: describe what the person is doing (hands/arms/body) in one short phrase.}
\prompttext{objects: list the main target objects the person is acting on / interacting with (not background clutter).}
\prompttext{summary: a short description of the scene in one sentence.}

\prompttext{{video frames}}
\end{mybox}
Each candidate model then receives the two structured captions, the relevant
label inventory, and this classification prompt:
\begin{mybox}[Video classification]
\footnotesize
\prompttext{You are a helpful AI assistant tasked with answering questions about videos
based on descriptions. }
\prompttext{Read the provided text carefully and return ONLY the
ID (e.g. c001, v003, o012) representing the correct answer.}
\medskip
\prompttext{Task: Identify the {activity/verb/object} from video descriptions.}
\prompttext{Determine the correct id using the structured descriptions below.}
\medskip
\prompttext{First-person view description: {ego_caption}}
\prompttext{Third-person view description: {exo_caption}}
\prompttext{Available {activity/verb/object}s:
{id}: {label}
...}
\medskip
\prompttext{Provide your answer as the ID (e.g., c001, v003, o012).}
\end{mybox}

\paragraph{Personalized dialogue preference.}
For MT-Bench, the original multi-turn user/assistant message sequence is
preserved; for Chatbot Arena, the original user message is preserved. No
additional fixed instruction is inserted before a candidate answers. Two
candidate responses are then compared by a judge conditioned on a sampled
persona:
\begin{mybox}[Persona preference judge]
\footnotesize
\prompttext{You are simulating the following user persona:
{persona}.}

\prompttext{As this persona, evaluate which assistant response you would personally
prefer.}
\medskip
\prompttext{[User Question]
{question}}

\prompttext{[Assistant 1]
{answer_1}}

\prompttext{[Assistant 2]
{answer_2}}
\medskip
\prompttext{Output ONLY one of the following tokens exactly:
1
2
Tie
Do not output anything else.}
\end{mybox}

\subsection{Router-Internal Templates}

The preceding templates construct the query--model response matrix used by
all routers. Most single-turn routers consume this matrix without another LLM
prompt. The following templates are used only by the routers that perform
additional language-model calls during routing or aggregation.

\paragraph{kNN-MultiRound and LLM-MultiRound.}
Both multi-round routers decompose a query, obtain sub-answers, and aggregate
them. kNN-MultiRound selects a model for each sub-query with nearest-neighbour
lookup, whereas LLM-MultiRound asks an LLM to choose the model. Their shared
decomposition template is:
\begin{mybox}[Multi-round decomposition]
\footnotesize
\prompttext{Given the query '{query}', decompose it into as many as 4 meaningful
sub-queries (minimum 1, maximum 4).}

\prompttext{Try to cover the full scope of the original query by breaking it down into
multiple specific and distinct sub-tasks whenever possible.}

\medskip
\prompttext{Aim for the maximum number of high-quality sub-queries without introducing
redundancy.}

\prompttext{Each sub-query should be clear, self-contained, and semantically coherent.}
\medskip
\prompttext{Output only the decomposed sub-queries, one per line.
Do not include any other text, explanations, or headers in your output.
Each line should contain only one sub-query.}
\end{mybox}
LLM-MultiRound appends the candidate names and descriptions, then replaces the
last three lines by the following routing constraint:
\begin{mybox}[Multi-round model selection]
\footnotesize

\prompttext{You will then be provided with descriptions of the following Large Language
Models (LLMs): {model_list}.}
\medskip
\prompttext{{model_descriptions}}
\medskip
\prompttext{For each sub-query, select the single LLM that is most likely to generate the
highest-quality response, regardless of cost or efficiency.}

\prompttext{Focus entirely on maximizing effectiveness and providing the most accurate
and relevant output. Base your decision strictly on the descriptions of the models.}
\medskip
\prompttext{Output only the decomposed sub-queries and the full name of the selected LLM
for each. Output exactly one sub-query and one LLM per line.}
\medskip
\prompttext{Each line must be formatted as follows:}

\prompttext{<sub-query>: <LLM name>}
\medskip
\prompttext{Use a colon ':' as the separator. Do not include any other text, explanations, or headers in your output.}

\end{mybox}
Each selected model is queried with:
\begin{mybox}[Multi-round specialist]
\footnotesize

\prompttext{You are a helpful assistant. You are participating in a multi-agent reasoning process, where a base model
delegates sub-questions to specialized models like you.}
\medskip
\prompttext{Your task is to do your absolute best to either:}

\prompttext{Answer the question directly, if possible, and provide a brief explanation;}

\prompttext{Offer helpful and relevant context, background knowledge, or insights related
to the question, even if you cannot fully answer it.}
\medskip
\prompttext{If you are completely unable to answer the question or provide any relevant or
helpful information, you must:}

\prompttext{Clearly state that you are unable to assist with this question.}

\prompttext{Explicitly instruct the base model to consult other LLMs for further
assistance.}
\prompttext{Keep your response clear, concise, and informative (preferably under 512
tokens).}

\prompttext{Stay strictly on-topic.}

\prompttext{Do not include irrelevant or generic content.}
\medskip
\prompttext{Here is the sub-question for you to assist with: {sub_query}}

\end{mybox}
For non-multiple-choice tasks, the aggregator receives:
\begin{mybox}[Multi-round aggregation]
\footnotesize

\prompttext{You are given a question along with auxiliary information, which consists of
several sub-questions derived from the original question and their respective
answers.}

\prompttext{Use this information to answer the original question if relevant, but make
your own reasoning step by step before arriving at the final answer.}
\medskip
\prompttext{Important: Your final answer MUST be clearly marked and enclosed within
<answer> and </answer> tags at the end of your response. No other part of the output should be inside these tags.}
\medskip
\prompttext{Auxiliary Information: {sub_query_and_response_pairs}}

\prompttext{Question: {query}}

\prompttext{Let's think step by step.}

\end{mybox}
For multiple-choice tasks, it instead receives:
\begin{mybox}[Multi-round multiple-choice aggregation]
\footnotesize

\prompttext{You are given a multiple-choice question and supporting sub-answers. Use the information only if helpful.}
\medskip
\prompttext{Question: {original_query}}

\prompttext{Supporting information: {sub_query_and_response_pairs}}
\medskip
\prompttext{Rules:}

\prompttext{Select exactly one option: A, B, C, D, or E.}

\prompttext{Output only the letter in <answer> tags.}

\prompttext{No explanation.}
\medskip
\prompttext{Format:}

\prompttext{<answer>A</answer>}

\end{mybox}

\paragraph{Router-R1.}
Router-R1 appends descriptions of the available models to the following
template. Its agent can request a specialist model within \texttt{<search>}
tags; the returned text is inserted as \texttt{<information>} in the next
round.
\begin{mybox}[Router-R1]
\footnotesize

\prompttext{Answer the given question.}

\prompttext{Every time you receive new information, you must first conduct reasoning
inside <think> ... </think>.}

\prompttext{After reasoning, if you find you lack some knowledge, you can call a
specialized LLM by writing a query inside
<search> LLM-Name:Your-Query </search>.}
\medskip
\prompttext{STRICT FORMAT RULES for <search>:}

\prompttext{Use an exact model name from: {candidate_model_names}.}

\prompttext{Replace Your-Query with a concrete question.}

\prompttext{Never copy model descriptions into <search>.}

\prompttext{Never output the placeholder format literally.}

\prompttext{Before each LLM call, reason about why external information is needed and
which model is appropriate.}

\prompttext{The returned response appears between <information> and </information>.}

\prompttext{Different models may be called multiple times.}
\medskip
\prompttext{{model_descriptions}}
\medskip
\prompttext{If no further external knowledge is needed, output the final answer inside
<answer> ... </answer>.}

\prompttext{Do not output empty answer tags.}
\medskip
\prompttext{Question: {question}}

\end{mybox}

\paragraph{CausalLM.}
The CausalLM router is trained to complete the selected model name after this
prefix:
\begin{mybox}[CausalLM]
\footnotesize
\prompttext{You are an intelligent router that selects the best Large Language Model (LLM)
for a given query.}
\medskip
\prompttext{Available LLMs: {model_1}, {model_2}, ...}

\prompttext{Based on the query content, complexity, and requirements, predict which LLM
would provide the best response.}
\medskip
\prompttext{Query: {query}}

\prompttext{Best LLM:}
\end{mybox}

\paragraph{AutoMix.}
AutoMix first asks a small model to answer with the following template:
\begin{mybox}[AutoMix answer generation]
\footnotesize
\prompttext{You are given a question. Answer the question as concisely as you can, using a
single phrase if possible.}
\medskip
\prompttext{Question: {question}}

\prompttext{Answer: The answer is'}
\end{mybox}
It uses the following few-shot verifier:
\begin{mybox}[AutoMix verifier]
\footnotesize
\prompttext{Question: Whose lost work was discovered in a dusty attic in 1980?}

\prompttext{AI Generated Answer: Shakespeare}

\prompttext{Instruction: Your task is to evaluate if the AI Generated Answer is correct,
based on the provided question. Provide the judgement and reasoning for each
case. Choose between Correct or Incorrect.}

\prompttext{Evaluation: The lost work of Shakespeare was discovered in 1980 in a dusty
attic.}

\prompttext{Verification Decision: The AI generated answer is Correct.}

\prompttext{---}

\prompttext{Question: In which month does the celestial event, the Pink Moon, occur?}

\prompttext{AI Generated Answer: July}

\prompttext{Instruction: Your task is to evaluate if the AI Generated Answer is correct,
based on the provided question. Provide the judgement and reasoning for each
case. Choose between Correct or Incorrect.}

\prompttext{Evaluation: The Pink Moon is unique to the month of April.}

\prompttext{Verification Decision: The AI generated answer is Incorrect.}

\prompttext{---}

\prompttext{Question: Who is believed to have painted the Mona Lisa in the early 16th
century?}

\prompttext{AI Generated Answer: Vincent van Gogh}

\prompttext{Instruction: Your task is to evaluate if the AI Generated Answer is correct,
based on the provided question. Provide the judgement and reasoning for each
case. Choose between Correct or Incorrect.}

\prompttext{Evaluation: The Mona Lisa was painted by Leonardo da Vinci in the early 16th
century.}

\prompttext{Verification Decision: The AI generated answer is Incorrect.}

\prompttext{---}

\prompttext{Question: How far away is the planet Kepler-442b?}

\prompttext{AI Generated Answer: 1,100 light-years}

\prompttext{Instruction: Your task is to evaluate if the AI Generated Answer is correct,
based on the provided question. Provide the judgement and reasoning for each
case. Choose between Correct or Incorrect.}

\prompttext{Evaluation: The Kepler-442b is located 1,100 light-years away.}

\prompttext{Verification Decision: The AI generated answer is Correct.}

\prompttext{---}

\prompttext{Question: {question}}

\prompttext{AI Generated Answer: {generated_answer}}

\prompttext{Instruction: Your task is to evaluate if the AI Generated Answer is correct,
based on the provided question. Provide the judgement and reasoning for each
case. Choose between Correct or Incorrect.}

\prompttext{Evaluation:}
\end{mybox}
The verifier's estimated correctness determines whether the query is escalated
to the larger model.

\paragraph{OpenClaw deployment router.}
When the deployed server uses prompt-based routing, it selects one model with:
\begin{mybox}[OpenClaw deployment router]
\footnotesize

\prompttext{You are an intelligent LLM router. Choose the most suitable model for the user's query.}
\medskip
\prompttext{Available models: {model_descriptions}}
\medskip
\prompttext{Rules:}

\prompttext{1. Simple greetings/daily chat -> cheaper models (8b, 9b size)}

\prompttext{2. Q&A/knowledge retrieval -> chatqa models}

\prompttext{3. Instruction following/structured output -> mistral models}

\prompttext{4. Code generation/technical questions -> nemotron or larger models}

\prompttext{5. Complex reasoning/deep analysis -> 70b or larger models}
\medskip
\prompttext{IMPORTANT: Only return the model name, nothing else!}
\medskip
\prompttext{Model names: {model_list}}

\prompttext{User query: {query}}

\end{mybox}
For image and video inputs, OpenClaw first uses the following preprocessing
prompts; the returned text is included in the router query.

\begin{mybox}[OpenClaw image and video preprocessing]
\footnotesize
\textbf{Image input:}

\prompttext{Describe this image concisely in 2--3 sentences.}

\textbf{Video input:}

\prompttext{Describe what you see in these video frames.}
\end{mybox}
\subsection{Multi-Agent Topology Templates}

The multi-agent topologies in \S\ref{app:mas} route every LLM-call node
independently from the prompt assigned to that node. Repeated nodes use the
same template and differ only in their indexed fields. The final node is shared
by all five topologies and is given after the topology-specific templates.

\paragraph{Star.}
The central planner decomposes the problem for \promptfield{n_actors} actors,
then consolidates their reports.

\begin{mybox}[Star planner]
\footnotesize
\prompttext{You are the central planner of a team of {n_actors} agents working on the problem
below. Decompose the problem into exactly {n_actors} subtasks, one per agent, so
that their combined results solve the problem. Output EXACTLY {n_actors} lines,
one subtask per line, no numbering or extra words.}
\medskip
\prompttext{Problem:
{user_query}}
\end{mybox}

\begin{mybox}[Star actor]
\footnotesize
\prompttext{You are agent {i+1} in a team. Solve the following subtask assigned by your
planner. Be concise and factual.}
\medskip
\prompttext{Original problem:
{user_query}}

\prompttext{Your subtask:
{subtask}}
\end{mybox}

\begin{mybox}[Star planner consolidation]
\footnotesize
\prompttext{You are the central planner. Your agents reported the results below. Consolidate
them into a single coherent analysis of the problem.}
\medskip
\prompttext{Problem:
{user_query}}

\prompttext{Agent reports:
Subtask: {subtask_1}
Result: {result_1}}

\prompttext{Subtask: {subtask_2}
Result: {result_2}}

\prompttext{...}
\end{mybox}

\paragraph{Tree.}
The root planner assigns two complementary sub-problems to team leads. Each
lead refines its branch into one task for a worker, after which the root planner
consolidates the two reports.

\begin{mybox}[Tree root planner]
\footnotesize
\prompttext{You are the top-level planner. Split the problem below into exactly 2
complementary sub-problems for your two subordinate team leads. Output EXACTLY
2 lines, one sub-problem per line, no numbering.}
\medskip
\prompttext{Problem:
{user_query}}
\end{mybox}

\begin{mybox}[Tree sub-planner]
\footnotesize
\prompttext{You are team lead {b+1}. Refine the sub-problem below into one concrete task
for your worker agent. Output ONE line only.}
\medskip
\prompttext{Original problem:
{user_query}}

\prompttext{Your sub-problem:
{branch_task}}
\end{mybox}

\begin{mybox}[Tree worker]
\footnotesize
\prompttext{You are a worker agent. Solve the task below concisely and factually.}
\medskip
\prompttext{Original problem:
{user_query}}

\prompttext{Your task:
{task}}
\end{mybox}

\begin{mybox}[Tree root consolidation]
\footnotesize
\prompttext{You are the top-level planner. Consolidate your two teams' reports into a
single coherent analysis of the problem.}
\medskip
\prompttext{Problem:
{user_query}}

\prompttext{Team reports:
Sub-problem: {branch_task_1}
Task: {task_1}
Result: {result_1}}

\prompttext{Sub-problem: {branch_task_2}
Task: {task_2}
Result: {result_2}}
\end{mybox}

\paragraph{Graph.}
Each of \promptfield{n_actors} agents first answers independently. In each
subsequent communication round, an agent receives the other agents' current
answers and revises its own answer.

\begin{mybox}[Graph actor: initial round]
\footnotesize
\prompttext{You are agent {i+1} of {n_actors} independent agents. Solve the problem below
on your own. Show brief reasoning, then your answer.}
\medskip
\prompttext{Problem:
{user_query}}
\end{mybox}

\begin{mybox}[Graph actor: revision round]
\footnotesize
\prompttext{You are agent {i+1} of {n_actors}. You communicated with all other agents;
their current answers are below. Reconsider and give your revised reasoning
and answer.}
\medskip
\prompttext{Problem:
{user_query}}

\prompttext{Your previous answer:
{answers[i]}}

\prompttext{Agent {j+1}'s current answer:
{answers[j]}}

\prompttext{...}
\end{mybox}

\paragraph{Chain.}
The first agent answers independently. Each later agent verifies and improves
the immediately preceding decision before passing its result onward.

\begin{mybox}[Chain: first agent]
\footnotesize
\prompttext{You are agent 1 in a chain of {n_agents} agents. Solve the problem below. Show
brief reasoning, then your answer. Your output will be passed to the next agent.}
\medskip
\prompttext{Problem:
{user_query}}
\end{mybox}

\begin{mybox}[Chain: later agent]
\footnotesize
\prompttext{You are agent {i+1} in a chain of {n_agents} agents. The previous agent passed
you their decision below. Verify it, fix any mistakes, and pass on your
improved reasoning and answer.}
\medskip
\prompttext{Problem:
{user_query}}

\prompttext{Previous agent's decision:
{decision}}
\end{mybox}

\paragraph{Plan-Exec-Sum.}
The GraphPlanner-style topology first decomposes the query into
\promptfield{width} atomic sub-queries. Each executor receives its sub-query
without an additional instruction, and the summarizer combines the results.

\begin{mybox}[Plan-Exec-Sum planner]
\footnotesize

\prompttext{You are a query decomposition assistant. Your task is to decompose the user's query into exactly {width} atomic and
independent sub-queries.}
\medskip
\prompttext{Keep each sub-query self-contained and non-overlapping.}

\prompttext{Prefer factual, directly answerable units.}

\prompttext{Output EXACTLY {width} lines, one sub-query per line, no numbering or extra
words.}

\medskip

\prompttext{User query: {user_query}}

\end{mybox}

\begin{mybox}[Plan-Exec-Sum executor]
\footnotesize
\prompttext{{sub_query}}
\end{mybox}

\begin{mybox}[Plan-Exec-Sum summarizer]
\footnotesize
\prompttext{You are a professional summarizer.
Summarize the following content into a concise, coherent paragraph without
bullet points.
Make it fluent and logically connected.}
\medskip
\prompttext{Content:
Sub-query: {sub_query_1}
Answer: {answer_1}}

\prompttext{Sub-query: {sub_query_2}
Answer: {answer_2}}

\prompttext{...}

\prompttext{Summary:}
\end{mybox}

\paragraph{Shared final node.}
Every topology uses the same final node. Its system instruction is restored
from the original task's \texttt{[System Instruction]} block, so the required
answer format remains task-specific. When there is team context, the final
node receives the following user prompt:

\begin{mybox}[Shared final node: user prompt with team context]
\footnotesize
\prompttext{{user_query}}

\prompttext{Below is supporting analysis from your team. Use it if helpful, but follow the
answer format required above.}

\prompttext{[Team Analysis]
{context}}
\end{mybox}

Without team context, the final node receives \promptfield{user_query} alone.

\section{The ComfyUI Visual Interface} \label{app:comfyui}

\method exposes its full routing pipeline as a graph on the ComfyUI canvas, where every node is a library component and every edge is an artifact that flows between components. Two input nodes supply the source benchmarks and the candidate pool $\mathcal{M}$, a data-engine node consumes them and emits the query--model matrix as a single edge, and each router node consumes that matrix and emits its evaluation. The matrix records the per-query performance and cost of every candidate, the supervision that every router in \S\ref{sec:general-router} is trained on. The graph traces the evaluation protocol of \S\ref{sec:protocol} from left to right, and the router nodes appear in menu groups named after the three router families, so the taxonomy of \S\ref{sec:general-router} is visible in the node menu. The canvas therefore renders the architecture of Figure~\ref{fig:library} as a diagram a user reads and edits, and replacing a router replaces one node while the data-engine and evaluation nodes stay in place, so the modularity of the library becomes visible on the canvas.

The data nodes realize the three stages of the data engine. The Select Datasets and Select LLMs nodes declare the query source and the candidate pool, and the Generate Data node runs query curation, response collection, and scoring in one call and writes the query--model matrix to a self-contained data directory. The router nodes cover every built-in router and are grouped in the menu by router family, and each router node reads its defaults from the same YAML configuration that the command line uses and renders every hyperparameter as a typed widget whose value, range, and options come from that file, so a canvas node and its scripted counterpart run one configuration. On execution a router node writes a runtime configuration, dispatches training and evaluation through the shared router registry, and returns a summary of the query count, the success count, the average performance, and the routing distribution over candidates. The Generate Data node hashes the selected datasets, candidate pool, and sample size into a metadata record and reuses an existing data directory when the record and its files are unchanged, which skips the costly response-collection stage on repeated runs.

\begin{figure}[H]
\centering
\includegraphics[width=\linewidth]{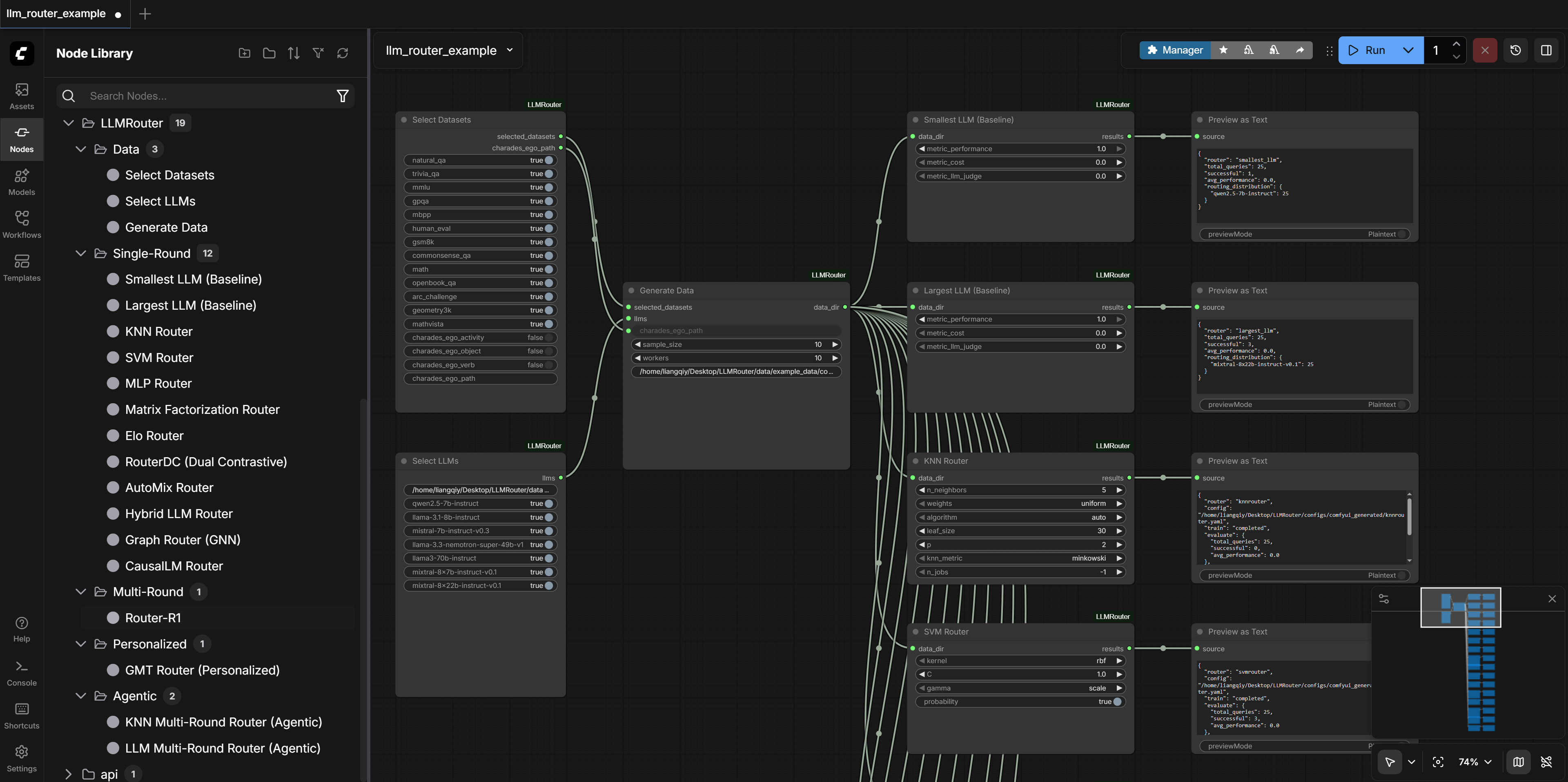}
\caption{\textbf{The ComfyUI interface of \method.} The source benchmarks and the candidate pool enter at the left, the data-engine node produces the query--model matrix, and each router node consumes the matrix and reports its evaluation, so the graph traces the routing pipeline from data construction to evaluation. Each router node exposes the hyperparameters of its library configuration as typed widgets.}
\label{fig:comfyui}
\end{figure}

\end{document}